\documentclass[10pt,twocolumn,letterpaper]{article}
\usepackage{amsmath,bm}
\usepackage{url}
\usepackage[pagenumbers]{cvpr} 

\usepackage[dvipsnames]{xcolor}

\definecolor{cvprblue}{rgb}{0.21,0.49,0.74}
\usepackage[pagebackref,breaklinks,colorlinks,citecolor=cvprblue]{hyperref}
\usepackage{indentfirst}
\usepackage{multirow}%

\def\paperID{4381} 
\def\confName{CVPR}
\def\confYear{2024}

\title{FaceChain-SuDe: Building Derived Class
to Inherit Category Attributes
for One-shot Subject-Driven Generation}

\newcommand*\samethanks[1][\value{footnote}]{\footnotemark[#1]}
\author{Pengchong Qiao$^{1,2}$\thanks{Equal contribution.}
\; Lei Shang$^{2}\samethanks$ \; Chang Liu$^{3}$\thanks{Corresponding author.} \; 
Baigui Sun$^{2}$ \; Xiangyang Ji$^{3}$ \; Jie Chen$^{1,4}$ \ \\
\normalsize{$^{1}$Peking University \quad $^{2}$Alibaba Group \quad $^{3}$Tsinghua University} \quad \normalsize{$^{4}$Pengcheng Laboratory} \\
{\tt\small pcqiao@stu.pku.edu.cn \quad \{sl172005, baigui.sbg\}@alibaba-inc.com} \\
{\tt\small \{liuchang2022, xyji\}@tsinghua.edu.cn} \quad \tt\small chenj@pcl.ac.cn
}

\begin{document}
\maketitle
\begin{abstract}
Subject-driven generation has garnered significant interest recently due to its ability to personalize text-to-image generation.
Typical works focus on learning the new subject's private attributes.
However, an important fact has not been taken seriously that a subject is not an isolated new concept but should be a specialization of a certain category in the pre-trained model.
This results in the subject failing to comprehensively inherit the attributes in its category, causing poor attribute-related generations.
In this paper, motivated by object-oriented programming, we model the subject as a derived class whose base class is its semantic category.
This modeling enables the subject to inherit public attributes from its category while learning its private attributes from the user-provided example.
Specifically, we propose a plug-and-play method, Subject-Derived regularization (SuDe).
It constructs the base-derived class modeling by constraining the subject-driven generated images to semantically belong to the subject's category.
Extensive experiments under three baselines and two backbones on various subjects show that our SuDe enables imaginative attribute-related generations while maintaining subject fidelity.
Codes will be open sourced soon at \href{https://github.com/modelscope/facechain}{FaceChain}.






\end{abstract}    
\vspace{-0.5em}
\section{Introduction}
\label{sec:intro}
Recently, with the fast development of text-to-image diffusion models~\cite{saharia2022photorealistic,ramesh2022hierarchical,nichol2022glide,rombach2022high}, people can easily use text prompts to generate high-quality, photorealistic, and imaginative images.
This gives people an outlook on AI painting in various fields such as game design, film shooting, etc.

\begin{figure}[htbp]
\centerline{\includegraphics[scale=0.65]{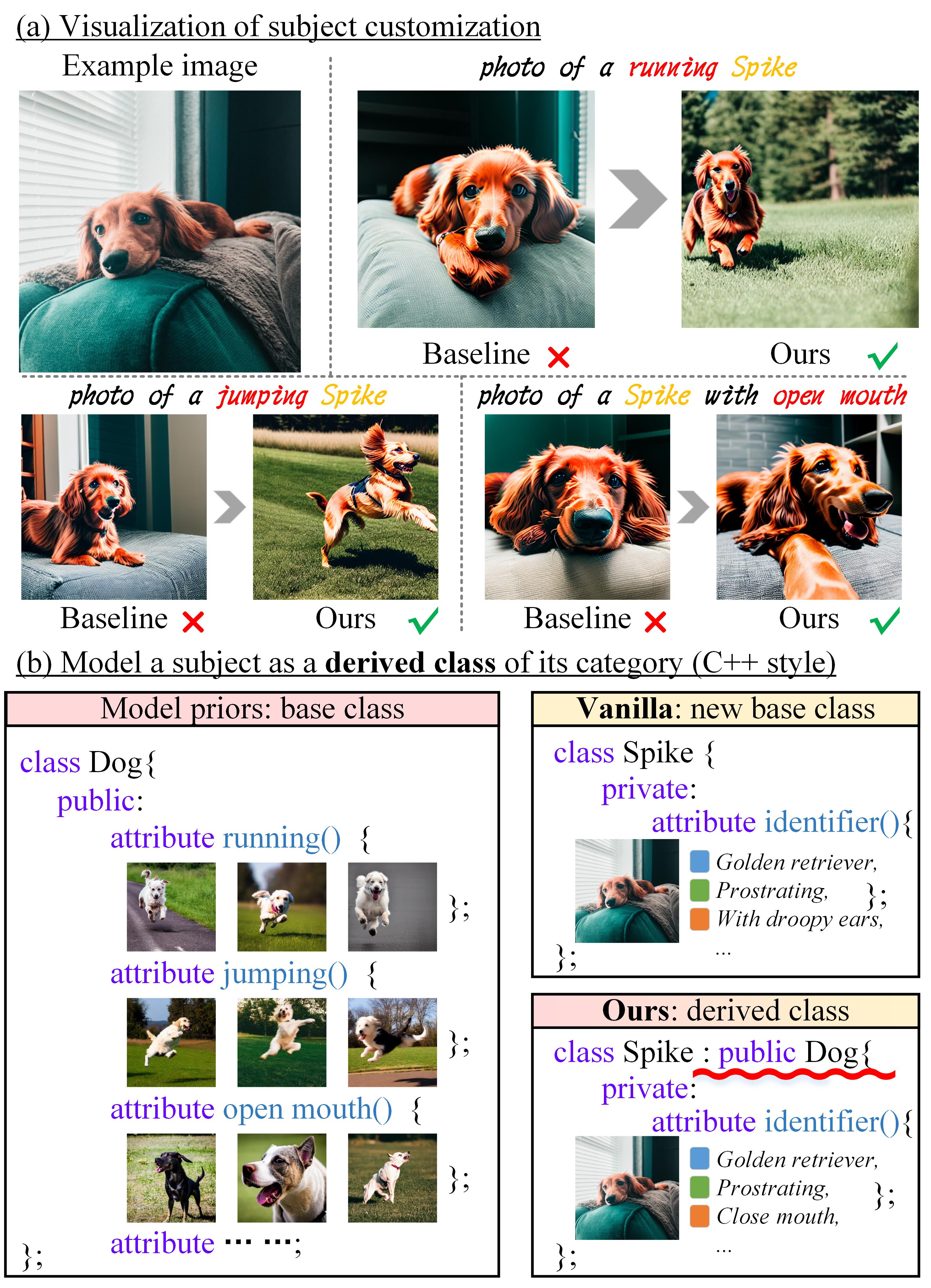}}
\caption{(a) The subject is a golden retriever `Spike', and the baseline is DreamBooth~\cite{ruiz2023dreambooth}.
The baseline's failure is because the example image cannot provide the needed attributes like `running'.
Our method tackles it by inheriting these attributes from the `Dog' category to `Spike'.
(b) We build `Spike' as a derived class of the base class `Dog'.
In this paper, we record the general properties of the base class from the pre-trained model as \textit{public attributes}, while subject-specific properties as \textit{private attributes}.
The part marked with a \textcolor{red}{red wavy line} is the `Inherit' syntax in C++~\cite{stroustrup1986overview}.
}
\label{fig: intro1}
\vspace{-1.5em}
\end{figure}

Among them, subject-driven generation is an interesting application that aims at customizing generation for a specific subject.
For example, something that interests you like pets, pendants, anime characters, etc.
These subjects are specific to each natural person (user) and do not exist in the large-scale training of pre-trained diffusion models.
To achieve this application, users need to provide a few example images to bind the subject with a special token (\{S$^{*}$\}), which could then be used to guide further customizations.

Existing methods can be classified into two types: offline ones and online ones.
The former~\cite{wei2023elite,ruiz2023hyperdreambooth} employs an offline trained encoder to directly encode the subject examples into text embedding, achieving high testing efficiency.
But the training of their encoders depends on an additional large-scale image dataset, and even the pixel-level annotations are also needed for better performances~\cite{wei2023elite}.
The latter~\cite{gal2022image, hao2023vico,kumari2023multi,ruiz2023dreambooth} adopts a test-time fine-tuning strategy to obtain the text embedding representing a specific subject.
Despite sacrificing testing efficiency, this kind of method eliminates reliance on additional data and is more convenient for application deployment.
Due to its flexibility, we focus on improving the online methods in this paper.

In deployment, the most user-friendly manner only requires users to upload one example image, called \textit{one-shot} subject-driven generation.
However, we find existing methods do not always perform satisfactorily in this challenging but valuable scene, especially for attribute-related prompts.
As shown in Fig.~\ref{fig: intro1} (a), the baseline method fails to make the `Spike' run, jump, or open its mouth, which are natural attributes of dogs.
Interestingly, the pre-trained model can generate these attributes for non-customized `Dogs'~\cite{saharia2022photorealistic,ramesh2022hierarchical,nichol2022glide,rombach2022high}.
From this, we infer that the failure in Fig.~\ref{fig: intro1} is because the single example image is not enough to provide the attributes required for customizing the subject, and these attributes cannot be automatically completed by the pre-trained model.
With the above considerations, we propose to tackle this problem by making the subject (`Spike') explicitly inherit these attributes from its semantic category (`Dog').
Specifically, motivated by the definitions in Object-Oriented Programming (OOP), we model the subject as a derived class of its category.
As shown in Fig.~\ref{fig: intro1} (b), the semantic category (`Dog') is viewed as a base class, containing public attributes provided by the pre-trained model.
The subject (`Spike') is modeled as a derived class of `Dog' to inherit its public attributes while learning private attributes from the user-provided example.
From the visualization in Fig.~\ref{fig: intro1} (a), our modeling significantly improves the baseline for attribute-related generations.

From the perspective of human understanding, the above modeling, i.e., subject (`Spike') is a derived class of its category (`Dog'), is a natural fact.
But it is unnatural for the generative model (e.g., diffusion model) since it has no prior concept of the subject `Spike'.
Therefore, to achieve this modeling,
we propose a \textbf{Subject Derivation regularization (SuDe)} to constrain that the generations of a subject could be classified into its corresponding semantic category.
Using the example above, generated images of `photo of a Spike' should have a high probability of belonging to `photo of a Dog'.
This regularization cannot be easily realized by adding a classifier since its semantics may misalign with that in the pre-trained diffusion model.
Thus, we propose to explicitly reveal the implicit classifier in the diffusion model to regularize the above classification.

Our SuDe is a plug-and-play method that can combine with existing subject-driven methods conveniently.
We evaluate this on three well-designed baselines, DreamBooth~\cite{ruiz2023dreambooth}, Custom Diffusion~\cite{kumari2023multi}, and ViCo~\cite{hao2023vico}.
Results show that our method can significantly improve attributes-related generations while maintaining subject fidelity.

\noindent Our main contributions are as follows:

\begin{itemize}
\item We provide a new perspective for subject-driven generation, that is, modeling a subject as a derived class of its semantic category, the base class.

\item We propose a subject-derived regularization (SuDe) to build the base-derived class relationship between a subject and its category with the implicit diffusion classifier.

\item Our SuDe can be conveniently combined with existing baselines and significantly improve attributes-related generations while keeping fidelity in a plug-and-play manner.

\end{itemize}

\section{Related Work}
\subsection{Object-Oriented Programming}
Object-Oriented Programming (OOP) is a programming paradigm with the concept of objects~\cite{rentsch1982object,wegner1990concepts,stroustrup1988object}, including four important definitions: class, attribute, derivation, and inheritance.
A \textit{class} is a template for creating objects containing some \textit{attributes}, which include public and private ones.
The former can be accessed outside the class, while the latter cannot.
\textit{Derivation} is to define a new class that belongs to an existing class, e.g., a new `Golden Retriever' class could be derived from the `Dog' class, where the former is called derived class and the latter is called base class.
\textit{Inheritance} means that the derived class should inherit some attributes of the base class, e.g., `Golden Retriever' should inherit attributes like `running' and `jumping' from `Dog'.

In this paper, we model the subject-driven generation as class derivation, where the subject is a derived class and its semantic category is the corresponding base class.
To adapt to this task, we use \textit{public attributes} to represent general properties like `running', and \textit{private attributes} to represent specific properties like the subject identifier.
The base class (category) contains public attributes provided by the pre-trained diffusion model and the derived class (subject) learns private attributes from the example image while inheriting its category's public attributes.



\subsection{Text-to-image generation}
Text-to-image generation aims to generate high-quality images with the guidance of the input text,
which is realized by combining generative models with image-text pre-trained models, e.g., CLIP~\cite{radford2021learning}.
From the perspective of generators, they can be roughly categorized into three groups: GAN-based, VAE-based, and Diffusion-based methods.
The GAN-based methods~\cite{reed2016generative,zhu2019dm,tao2022df,xu2018attngan,crowson2022vqgan} employ the Generative Adversarial Network as the generator and perform well on structural images like human faces.
But they struggle in complex scenes with varied components.
The VAE-based methods~\cite{chang2023muse,ding2021cogview,gafni2022make,ramesh2021zero} generate images with Variational Auto-encoder, which can synthesize diverse images but sometimes cannot match the texts well.
Recently, Diffusion-based methods~\cite{ding2022cogview2,nichol2022glide,ramesh2022hierarchical,rombach2022high,saharia2022photorealistic,balaji2022ediff} obtain SOTA performances and can generate photo-realistic images according to the text prompts.
In this paper, we focus on deploying the pre-trained text-to-image diffusion models into the application of subject-customization.





\subsection{Subject-driven generation}
Given a specific subject, subject-driven generation aims to generate new images of this subject with text guidance.
Pioneer works can be divided into two types according to training strategies, the offline and the online ones.
Offline methods~\cite{wei2023elite,ruiz2023hyperdreambooth,chen2023disenbooth,chen2023subject} directly encode the example image of the subject into text embeddings, for which they need to train an additional encoder.
Though high testing efficiency, they are of high cost since a large-scale dataset is needed for offline training.
Online methods~\cite{gal2022image, hao2023vico,kumari2023multi,ruiz2023dreambooth,tewel2023key} learn a new subject in a test-time tuning manner.
They represent the subject with a specific token `\{S$^{*}$\}' by fine-tuning the pre-trained model in several epochs.
Despite sacrificing some test efficiency, they don't need additional datasets and networks.
But for the most user-friendly one-shot scene, these methods cannot customize attribute-related generations well.
To this end, we propose to build the subject as a derived class of its category to inherit public attributes while learning private attributes.
Some previous works~\cite{ruiz2023dreambooth,kumari2023multi} partly consider this problem by prompt engineering, but we show our SuDe is more satisfactory, as in sec.~\ref{sec: prompt}.




\begin{figure*}[htbp]
\centerline{\includegraphics[scale=0.8]{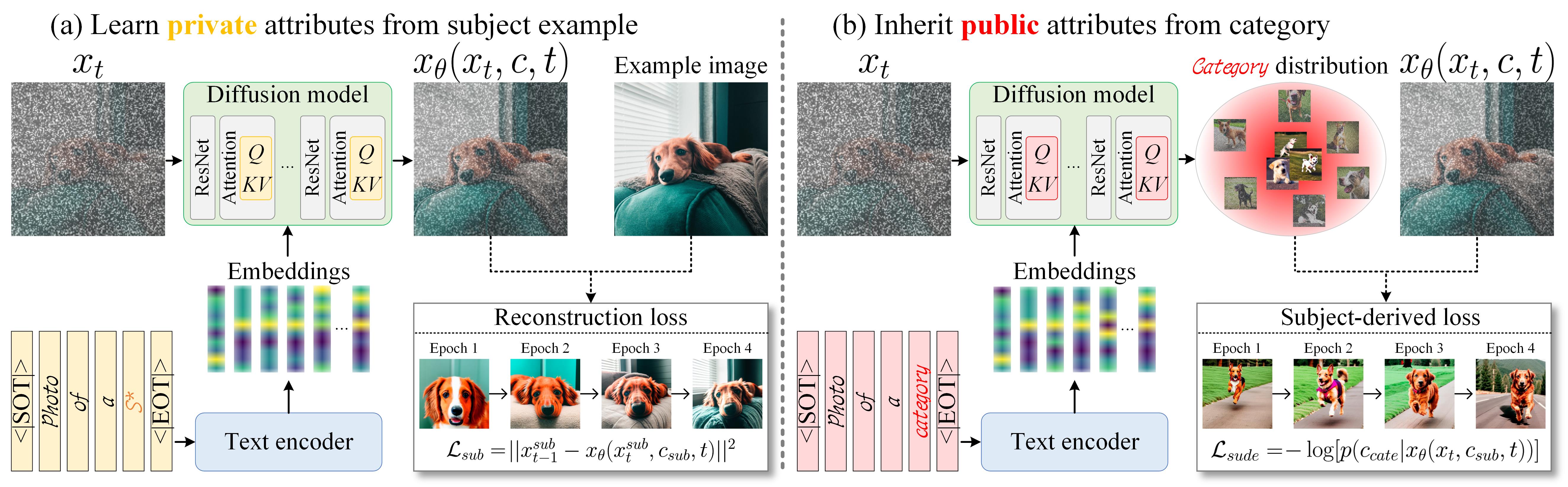}}
\caption{\textbf{The pipeline of SuDe.} (a)  Learn private attributes by reconstructing the subject example with the $\mathcal{L}_{sub}$ in Eq.~\ref{eq: subject loss}. (b) Inherit public attributes by constraining the subject-driven $\bm{x}_{t-1}$ semantically belongs to its category (e.g., dog), with the $\mathcal{L}_{sude}$ in Eq.~\ref{eq: sude loss}.
}
\label{fig: framework}
\vspace{-1.0em}
\end{figure*}

\section{Method}
\subsection{Preliminaries}

\subsubsection{Text-to-image diffusion models}\label{sec: diffusion model}
Diffusion models~\cite{ho2020denoising,sohl2015deep} approximate real data distribution by restoring images from Gaussian noise.
They use a forward process gradually adding noise $\bm{\epsilon} \sim \mathcal{N}(\mathbf{0},\mathbf{I})$ on the clear image (or its latent code) $\bm{x}_0$ to obtain a series of noisy variables $\bm{x}_1$ to $\bm{x}_T$, where $T$ usually equals 1000, as:
\begin{equation}
  \begin{aligned}
\bm{x}_t = \sqrt{\alpha_{t}} \bm{x}_0+\sqrt{1-\alpha_t} \bm{\epsilon},
  \end{aligned} \label{eq: xt}
\end{equation}
where $\alpha_t$ is a $t$-related variable that controls the noise schedule.
In text-to-image generation, a generated image is guided by a text description $\bm{P}$.
Given a noisy variable $\bm{x}_t$ at step $t$, the model is trained to denoise the $\bm{x}_t$ gradually as:
\begin{equation}
  \begin{aligned}
\mathbb{E}_{\bm{x},\bm{c},\bm{\epsilon},t}[w_{t}|| \bm{x}_{t-1} - x_{\theta}(\bm{x}_{t}, \bm{c}, t)||^{2}],
  \end{aligned} \label{eq: diffusion loss eps}
\end{equation}
where $x_{\theta}$ is the model prediction, $w_t$ is the loss weight at step $t$, $\bm{c} = \Gamma(\bm{P})$ is the embedding of text prompt, and the $\Gamma(\cdot)$ is a pre-trained text encoder, such as BERT~\cite{kenton2019bert}.
In our experiments, we use Stable Diffusion~\cite{2022sd} built on LDM~\cite{rombach2022high} with the CLIP~\cite{radford2021learning} text encoder as our backbone model.

\subsubsection{Subject-driven finetuning}\label{sec: subject-dirven base}
\textbf{Overview:}
The core of the subject-driven generation is to implant the new concept of a subject into the pre-trained diffusion model.
Existing works~\cite{gal2022image, hao2023vico,ruiz2023dreambooth,kumari2023multi,zhang2023prospect} realize this via finetuning partial or all parameters of the diffusion model, or text embeddings, or adapters, by:
\begin{equation}
  \begin{aligned}
\mathcal{L}_{sub} = ||\bm{x}_{t-1} - x_{\theta}(\bm{x}_{t}, \bm{c}_{sub}, t)||^{2},
  \end{aligned} \label{eq: subject loss}
\end{equation}
where the $\bm{x}_{t-1}$ here is the noised user-provided example at step $t-1$, $\bm{c}_{sub}$ is the embedding of subject prompt (e.g., `photo of a \{S$^{*}$\}'). The `\{S$^{*}$\}' represents the subject name.


\textbf{Motivation:} 
With Eq.~\ref{eq: subject loss} above, existing methods can learn the specific attributes of a subject.
However, the attributes in the user-provided single example are not enough for imaginative customizations.
Existing methods haven't made designs to address this issue, only relying on the pre-trained diffusion model to fill in the missing attributes automatically.
But we find this is not satisfactory enough, e.g., in Fig.~\ref{fig: intro1}, baselines fail to customize the subject `Spike' dog to `running' and `jumping'.
To this end, we propose to model a subject as a derived class of its semantic category, the base class.
This helps the subject inherit the public attributes of its category while learning its private attributes and thus improves attribute-related generation while keeping subject fidelity.
Specifically, as shown in Fig.~\ref{fig: framework} (a), the private attributes are captured by reconstructing the subject example.
And the public attributes are inherited via encouraging the subject prompt (\{$S^{*}$\}) guided $\bm{x}_{t-1}$ to semantically belong to its category (e.g., `Dog'), as Fig.~\ref{fig: framework} (b).

\subsection{Subject Derivation Regularization}
Derived class is a definition in object-oriented programming, not a proposition.
Hence there is no sufficient condition that can be directly used to constrain a subject to be a derived class of its category.
However, according to the definition of derivation, there is naturally a necessary condition: a derived class should be a subclass of its base class.
We find that constraining this necessary condition is very effective for helping a subject to inherit the attributes of its category.
Specifically, we regularize the subject-driven generated images to belong to the subject's category as:
\begin{equation}
  \begin{aligned}
  \mathcal{L}_{sude} = -\log[p(\bm{c}_{cate}|x_{\theta}(\bm{x}_{t}, \bm{c}_{sub}, t))],
  \end{aligned} \label{eq: sude loss}
\end{equation}
where $\bm{c}_{cate}$ and $\bm{c}_{sub}$ are conditions of category and subject.
The Eq.~\ref{eq: sude loss} builds a subject as a derived class well for two reasons:
(1) The attributes of a category are reflected in its embedding $\bm{c}_{cate}$, most of which are public ones that should be inherited.
This is because the embedding is obtained by a pre-trained large language model (LLM)~\cite{kenton2019bert}, which mainly involves general attributes in its training.
(2) As analyzed in Sec.~\ref{sec: analysis}, optimizing $\mathcal{L}_{sude}$ combined with the Eq.~\ref{eq: subject loss} is equivalent to increasing $p(\bm{x}_{t-1} | \bm{x}_{t}, \bm{c}_{sub}, \bm{c}_{cate})$, which means generating a sample with the conditions of both $\bm{c}_{sub}$ (private attributes) and $\bm{c}_{cate}$ (public attributes).
Though the form is simple, Eq.~\ref{eq: sude loss} cannot be directly optimized.
In the following, we describe how to compute it in Sec.~\ref{sec: Reveal Implicit Classifier}, and a necessary strategy to prevent training crashes in Sec.~\ref{sec: Loss Value Truncate}.

\subsubsection{Subject Derivation Loss}\label{sec: Reveal Implicit Classifier}
The probability in Eq.~\ref{eq: sude loss} cannot be easily obtained by an additional classifier since its semantics may misalign with that in the pre-trained diffusion model.
To ensure semantics alignment, we propose to reveal the implicit classifier in the diffusion model itself.
With the Bayes' theorem~\cite{joyce2003bayes}:
\begin{equation}
  \begin{aligned}
  p(\bm{c}_{cate}|x_{\theta}(\bm{x}_{t}, \bm{c}_{sub}, t)) = C_{t} \cdot \frac{p(x_{\theta}(\bm{x}_{t}, \bm{c}_{sub}, t)|\bm{x}_{t},\bm{c}_{cate})}{p(x_{\theta}(\bm{x}_{t}, \bm{c}_{sub}, t)|\bm{x}_{t})},
  \end{aligned} \label{eq: reveal classifier}
\end{equation}
where the $C_t = p(\bm{c}_{cate}|\bm{x}_{t})$ is unrelated to $t-1$, thus can be ignored in backpropagation.
In the Stable Diffusion~\cite{2022sd}, predictions of adjacent steps (i.e., $t-1$ and $t$) are designed as a conditional Gaussian distribution:
\begin{equation}
  \begin{aligned}
&p(\bm{x}_{t-1}|\bm{x}_{t},\bm{c}) \sim \mathcal{N}(\bm{x}_{t-1};x_{\theta}(\bm{x}_{t}, \bm{c}, t), \sigma^{2}_{t} \mathbf{I}) \\
&\propto exp({-||\bm{x}_{t-1}-x_{\theta}(\bm{x}_t,\bm{c}, t)||^{2}/2\bm\sigma^{2}_{t}}),
  \end{aligned} \label{eq: normal distribution}
\end{equation}
where the mean value is the prediction at step $t$ and the standard deviation is a function of $t$.
From Eq.~\ref{eq: reveal classifier} and ~\ref{eq: normal distribution}, we can convert Eq.~\ref{eq: sude loss} into a computable form:
\begin{equation}
  \begin{aligned}
  \mathcal{L}_{sude} &= \frac{1}{2\bm{\sigma}^{2}_t} [||x_{\theta}(\bm{x}_{t}, \bm{c}_{sub}, t) - x_{\Bar{\theta}}(\bm{x}_t, \bm{c}_{cate}, t)||^{2} \\
  &-||x_{\theta}(\bm{x}_{t}, \bm{c}_{sub}, t)- x_{\Bar{\theta}}(\bm{x}_t, t)||^{2}],
  \end{aligned} \label{eq: sude loss p form}
\end{equation}
where the $x_{\Bar{\theta}}(\bm{x}_{t}, \bm{c}_{cate}, t)$ is the prediction conditioned on $\bm{c}_{cate}$, the $x_{\Bar{\theta}}(\bm{x}_t, t)$ is the unconditioned prediction.
The $\Bar{\theta}$ means detached in training, indicating that only the $x_{\theta}(\bm{x}_{t}, \bm{c}_{sub}, t)$ is gradient passable, and
the $x_{\Bar{\theta}}(\bm{x}_t, \bm{c}_{cate}, t)$ and $x_{\Bar{\theta}}(\bm{x}_t, t)$ are gradient truncated.
This is because they are priors in the pre-trained model that we want to reserve.


\begin{figure*}[htbp]
\centerline{\includegraphics[scale=0.6]{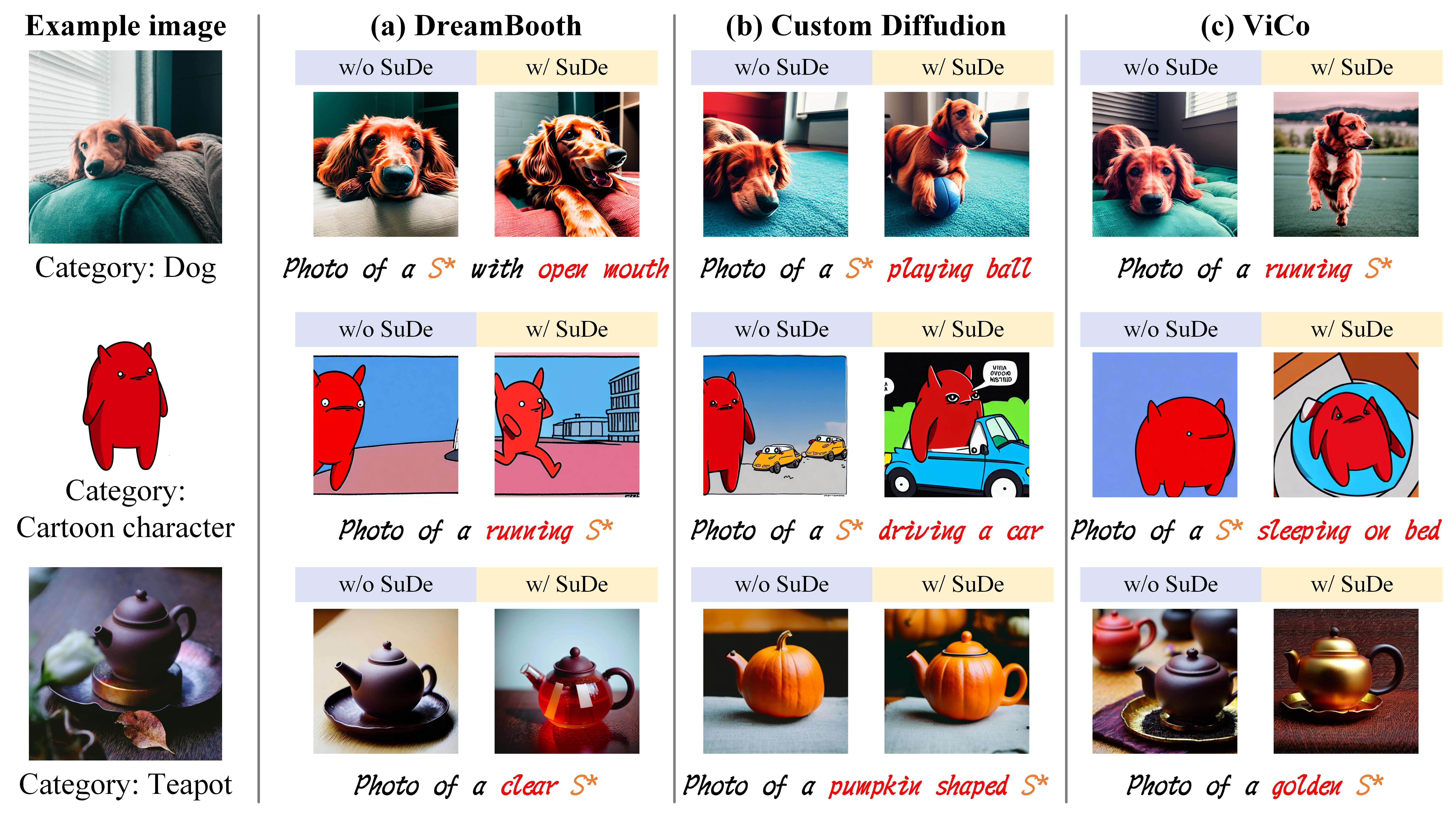}}
\caption{(a), (b), and (c) are generated images using DreamBooth~\cite{ruiz2023dreambooth}, Custom Diffusion~\cite{kumari2023multi}, and ViCo~\cite{hao2023vico} as the baselines, respectively.
Results are obtained using the DDIM~\cite{song2020denoising} sampler with 100 steps.
In prompts, we mark the subject token in \textcolor[RGB]{237,125,49}{orange} and attributes in  \textcolor{red}{red}.
}
\label{fig: visual result 1}
\vspace{-1.0em}
\end{figure*}

\subsubsection{Loss Truncation}\label{sec: Loss Value Truncate}
Optimizing Eq.~\ref{eq: sude loss} will leads the $p(\bm{c}_{cate}|x_{\theta}(\bm{x}_{t}, \bm{c}_{sub}, t))$ to increase until close to 1.
However, this term represents the classification probability of a noisy image at step $t-1$.
It should not be close to 1 due to the influence of noise.
Therefore, we propose to provide a threshold to truncate $\mathcal{L}_{sude}$.
Specifically, for generations conditioned on $\bm{c}_{cate}$, their probability of belonging to $\bm{c}_{cate}$ can be used as a reference.
It represents the proper classification probability of noisy images at step $t-1$.
Hence, we use the negative log-likelihood of this probability as the threshold $\tau$, which can be computed by replacing the $\bm{c}_{sub}$ with $\bm{c}_{cate}$ in Eq.~\ref{eq: sude loss p form}:
\begin{equation}
  \begin{aligned}
  \tau_{t} &= -\log[p(\bm{c}_{cate}|x_{\theta}(\bm{x}_{t}, \bm{c}_{cate}, t))]\\
  &=-\frac{1}{2\bm{\sigma}^{2}_t}||x_{\Bar{\theta}}(\bm{x}_{t}, \bm{c}_{cate}, t)- x_{\Bar{\theta}}(\bm{x}_t, t)||^{2}.
  \end{aligned} \label{eq: threshold}
\end{equation}
The Eq.~\ref{eq: threshold} represents the lower bound of $\mathcal{L}_{sude}$ at step $t$.
When the loss value is less than or equal to $\mathcal{L}_{sude}$, optimization should stop.
Thus, we truncate $\mathcal{L}_{sude}$ as:
\begin{equation}
\mathcal{L}_{sude} = \lambda_{\tau} \cdot \mathcal{L}_{sude}, ~~~\lambda_{\tau} = 
\left\{
  \begin{aligned}
  &0, ~~~~\mathcal{L}_{sude} \leq \tau_{t} \\
  &1,  ~~~~else.
  \end{aligned} \label{eq: threshold loss}
\right.
\end{equation}
In practice, this truncation is important for maintaining training stability.
Details are provided in Sec.~\ref{sec: ablation trunction}.


\subsection{Overall Optimization Objective}
Our method only introduces a new loss function $\mathcal{L}_{sude}$, thus it can be conveniently implanted into existing pipelines in a plug-and-play manner as:
\begin{equation}
  \begin{aligned}
  \mathcal{L} = \mathbb{E}_{\bm{x},\bm{c},\bm{\epsilon},t} [  \mathcal{L}_{sub} + w_{s} \mathcal{L}_{sude} + w_{r} \mathcal{L}_{reg}],
  \end{aligned} \label{eq: total loss}
\end{equation}
where $\mathcal{L}_{sub}$ is the reconstruction loss to learn the subject's private attributes as described in Eq.~\ref{eq: subject loss}.
The $\mathcal{L}_{reg}$ is a regularization loss usually used to prevent the model from overfitting to the subject example.
Commonly, it is not relevant to $\bm{c}_{sub}$ and has flexible definitions~\cite{ruiz2023dreambooth,hao2023vico} in various baselines.
The $w_{s}$ and $w_{r}$ are used to control loss weights.
In practice, we keep the $\mathcal{L}_{sub}$, $\mathcal{L}_{reg}$ follow baselines, only changing the training process by adding our $\mathcal{L}_{sude}$.





\begin{table*}[htbp]
\setlength{\tabcolsep}{1.6mm}
  \centering
  \caption{\textbf{Quantitative results.} These results are average on 4 generated images for each prompt with a DDIM~\cite{song2020denoising} sampler with 50 steps.
  The $^{\dagger}$ means performances obtained with a flexible $w_s$.
  The improvements our SuDe brought on the baseline are marked in red.
  }\label{tab: Quantitative Results}
\begin{tabular}{l|cccc|cccc}
\hline
\multirow{2}{*}{Method}  & \multicolumn{4}{c|}{Results on Stable diffusion v1.4 (\%)} & \multicolumn{4}{c}{Results on Stable diffusion v1.5 (\%)} \\ \cline{2-9} 
                         & CLIP-I    & DINO-I   & CLIP-T    & BLIP-T    & CLIP-I    & DINO-I    & CLIP-T   & BLIP-T   \\ \hline
ViCo~\cite{hao2023vico}                     & 75.4     & 53.5     & 27.1     & 39.1     & 78.5      & 55.7      & 28.5      & 40.7     \\
ViCo w/ SuDe             & 76.1     & 56.8     & 29.7 (\textcolor{red}{+2.6})     & 43.3 (\textcolor{red}{+4.2})     & 78.2      & 59.4      & 29.6 (\textcolor{red}{+1.1})     & 43.3 (\textcolor{red}{+2.6})    \\
ViCo w/ SuDe$^{\dagger}$             & 75.8     & 57.5     & 30.3 (\textcolor{red}{+3.2})     & 44.4 (\textcolor{red}{+5.3})     & 77.3      & 58.4      & 30.2 (\textcolor{red}{+1.7})      & 44.6 (\textcolor{red}{+3.9})    \\ \hline
Custom Diffusion~
\cite{kumari2023multi}         & 76.5     & 59.6     & 30.1     & 45.2     & 76.5      & 59.8      & 30.0      & 44.6     \\
Custom Diffusion w/ SuDe & 76.3     & 59.1     & 30.4 (\textcolor{red}{+0.3})     & 46.1 (\textcolor{red}{+0.9})     & 76.0      & 60.0      & 30.3 (\textcolor{red}{+0.3})      & 46.6 (\textcolor{red}{+2.0})    \\ 
Custom Diffusion w/ SuDe$^{\dagger}$ & 76.4     & 59.7     & 30.5 (\textcolor{red}{+0.4})     & \textbf{46.3} (\textcolor{red}{+1.1})     & 76.2      & 60.3      & \textbf{30.3} (\textcolor{red}{+0.3})      & \textbf{46.9} (\textcolor{red}{+2.3})     \\ \hline                 
DreamBooth~\cite{ruiz2023dreambooth}                & 77.4     & 59.7     & 29.0     & 42.1     & \textbf{79.5}      & \textbf{64.5}      & 29.0      & 41.8     \\
DreamBooth w/ SuDe       & \textbf{77.4}     & \textbf{59.9}     & 29.5 (\textcolor{red}{+0.5})     & 43.3 (\textcolor{red}{+1.2})     & 78.8      & 63.3      & 29.7 (\textcolor{red}{+0.7})      & 43.3 (\textcolor{red}{+1.5})    \\ 
DreamBooth w/ SuDe$^{\dagger}$       & 77.1     & 59.7     & \textbf{30.5} (\textcolor{red}{+1.5})     & 45.3 (\textcolor{red}{+3.2})     & 78.8      & 64.0      & 29.9 (\textcolor{red}{+0.9})     & 43.8 (\textcolor{red}{+2.0})     \\ \hline
\end{tabular}
\vspace{-1.0em}
\end{table*}

\section{Theoretical Analysis} \label{sec: analysis}
Here we analyze that SuDe works well since it models the $p(\bm{x}_{t-1} | \bm{x}_{t}, \bm{c}_{sub}, \bm{c}_{cate})$.
According to Eq.~\ref{eq: subject loss},~\ref{eq: sude loss} and DDPM~\cite{ho2020denoising}, we can express $\mathcal{L}_{sub}$ and $\mathcal{L}_{sude}$ as:
\begin{equation}
  \begin{aligned}
&\mathcal{L}_{sub} = -\log[p(\bm{x}_{t-1}|\bm{x}_{t}, \bm{c}_{sub})], \\
&\mathcal{L}_{sude} = -\log[p(\bm{c}_{cate}|\bm{x}_{t-1}, \bm{c}_{sub})].
  \end{aligned} \label{eq: analysis 1}
\end{equation}
Here we first simplify the $w_{s}$ to 1 for easy understanding:
\begin{equation}
  \begin{aligned}
  &\mathcal{L}_{sub} + \mathcal{L}_{sude} 
  = -\log[p(\bm{x}_{t-1}|\bm{x}_{t}, \bm{c}_{sub}) \cdot p(\bm{c}_{cate}|\bm{x}_{t-1}, \bm{c}_{sub})] \\
  &=-\log[ p(\bm{x}_{t-1}|\bm{x}_{t},\bm{c}_{sub}, \bm{c}_{cate}) \cdot p(\bm{c}_{cate}|\bm{x}_{t}, \bm{c}_{sub}) ] \\
  &=-\log[ p(\bm{x}_{t-1}|\bm{x}_{t},\bm{c}_{sub}, \bm{c}_{cate})] + S_{t},
  \end{aligned} \label{eq: analysis 2}
\end{equation}
where $S_{t} = -\log[p(\bm{c}_{cate}|\bm{x}_{t}, \bm{c}_{sub})]$ is unrelated to $t-1$.
Form this Eq.~\ref{eq: analysis 2}, we find that our method models the distribution of $p(\bm{x}_{t-1} | \bm{x}_{t}, \bm{c}_{sub}, \bm{c}_{cate})$, which takes both $\bm{c}_{sub}$ and $\bm{c}_{cate}$ as conditions, thus could generate images with private attributes from $\bm{c}_{sub}$ and public attributes from $\bm{c}_{cate}$.

In practice, $w_{s}$ is a changed hyperparameter on various baselines.
This does not change the above conclusion since:
\begin{equation}
  \begin{aligned}
  &w_s \cdot \mathcal{L}_{sude} = -\log[p^{w_{s}}(\bm{c}_{cate}|\bm{x}_{t-1}, \bm{c}_{sub})], \\
  & p^{w_{s}}(\bm{c}_{cate}|\bm{x}_{t-1}, \bm{c}_{sub}) \propto p(\bm{c}_{cate}|\bm{x}_{t-1}, \bm{c}_{sub}),
  \end{aligned} \label{eq: analysis 3}
\end{equation}
where the $a \propto b$ means $a$ is positively related to $b$.
Based on Eq.~\ref{eq: analysis 3}, we can see that the $\mathcal{L}_{sub} + w_s \mathcal{L}_{sude}$ is positively related to $-\log[ p(\bm{x}_{t-1}|\bm{x}_{t},\bm{c}_{sub}, \bm{c}_{cate})]$.
This means that optimizing our $\mathcal{L}_{sude}$ with $\mathcal{L}_{sub}$ can still increase $p(\bm{x}_{t-1}|\bm{x}_{t},\bm{c}_{sub}, \bm{c}_{cate})$ when $w_s$ is not equal to 1.

\section{Experiments}
\subsection{Implementation Details} \label{sec: Implementation Details}
\textbf{Frameworks:}
We evaluate that our SuDe works well in a plug-and-play manner on three well-designed frameworks, DreamBooth~\cite{ruiz2023dreambooth}, Custom Diffusion~\cite{kumari2023multi}, and ViCo~\cite{hao2023vico} under two backbones, Stable-diffusion v1.4 (SD-v1.4) and Stable-diffusion v1.5 (SD-v1.5)~\cite{2022sd}.
In practice, we keep all designs and hyperparameters of the baseline unchanged and only add our $\mathcal{L}_{sude}$ to the training loss.
For the hyperparameter $w_s$, since these baselines have various training paradigms (e.g., optimizable parameters, learning rates, etc), it's hard to find a fixed $w_s$ for all these baselines.
We set it to 0.4 on DreamBooth, 1.5 on ViCo, and 2.0 on Custom Diffusion.
A noteworthy point is that users can adjust $w_s$ according to different subjects in practical applications.
This comes at a very small cost because our SuDe is a plugin for test-time tuning baselines, which are of high efficiency (e.g., $\sim$ 7 min for ViCo on a single 3090 GPU).

\textbf{Dataset:}
For quantitative experiments, we use the DreamBench dataset provided by DreamBooth~\cite{ruiz2023dreambooth}, containing 30 subjects from 15 categories, where each subject has 5 example images.
Since we focus on one-shot customization here, we only use one example image (numbered `00.jpg') in all our experiments.
In previous works, their most collected prompts are attribute-unrelated, such as `photo of a \{S$^{*}$\} in beach/snow/forest/...', only changing the image background.
To better study the effectiveness of our method, we collect 5 attribute-related prompts for each subject.
Examples are like `photo of a \textit{running} \{S$^{*}$\}' (for dog), `photo of a \textit{burning} \{S$^{*}$\}' (for candle).
Moreover, various baselines have their unique prompt templates.
Specifically, for ViCo, its template is `photo of a \{S$^{*}$\}', while for DreamBooth and Custom Diffusion, the template is `photo of a \{S$^{*}$\} [category]'.
In practice, we use the default template of various baselines.
In this paper, for the convenience of writing, we uniformly record \{S$^{*}$\} and \{S$^{*}$\} [category] as \{S$^{*}$\}.
Besides, we also show other qualitative examples in appendix, which are collected from Unsplash~\cite{Unsplash}.

\textbf{Metrics:}
For the subject-driven generation task, two important aspects are \textit{subject fidelity} and \textit{text alignment}.
For the first aspect, we refer to previous works and use DINO-I and CLIP-I as the metrics.
They are the average pairwise cosine similarity between DINO~\cite{caron2021emerging} (or CLIP~\cite{radford2021learning}) embeddings of generated and real images.
As noted in~\cite{ruiz2023dreambooth, hao2023vico}, the DINO-I is better at reflecting fidelity than CLIP-I since DINO can capture differences between subjects of the same category.
For the second aspect, we refer to previous works that use CLIP-T as the metric, which is the average cosine similarity between CLIP~\cite{radford2021learning} embeddings of prompts and generated images.
Additionally, we propose a new metric to evaluate the text alignment about attributes, abbreviated as \textit{attribute alignment}.
This cannot be reflected by CLIP-T since CLIP is only coarsely trained at the classification level, being insensitive to attributes like actions and materials.
Specifically, we use BLIP-T, the average cosine similarity between BLIP~\cite{li2022blip} embeddings of prompts and generated images.
It can measure the attribute alignment better since the BLIP is trained to handle the image caption task.



\begin{figure*}[htbp]
\centerline{\includegraphics[scale=0.93]{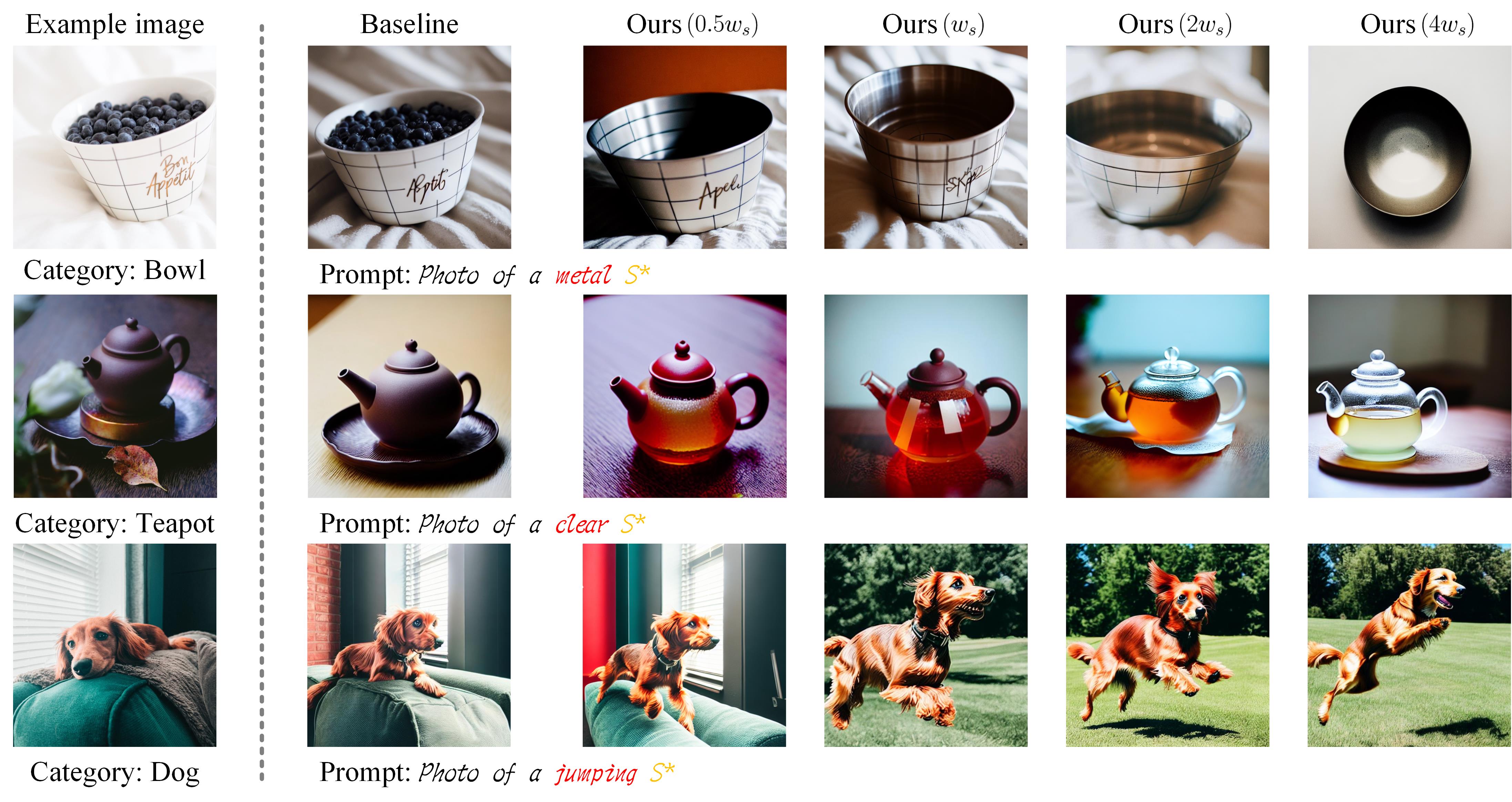}}
\vspace{-0.5em}
\caption{\textbf{Visual comparisons by using different values of $w_s$}. Results are from DreamBooth w/ SuDe, where the default $w_s$ is 0.4.
}
\label{fig: ablation w}
\vspace{-1.0em}
\end{figure*}

\subsection{Qualitative Results} \label{sec: Qualitative Results}
Here, we visualize the generated images on three baselines with and without our method in Fig.~\ref{fig: visual result 1}.

\textbf{Attribute alignment:}
Qualitatively, we see that generations with our SuDe align the attribute-related texts better.
For example, in the 1st row, Custom Diffusion cannot make the dog \textit{\textbf{playing ball}}, in the 2nd row, DreamBooth cannot let the cartoon character \textit{\textbf{running}}, and in the 3rd row, ViCo cannot give the teapot a \textit{\textbf{golden material}}.
In contrast, after combining with our SuDe, their generations can reflect these attributes well.
This is because our SuDe helps each subject inherit the public attributes in its semantic category.


\begin{figure}[htbp]
\centerline{\includegraphics[scale=0.54]{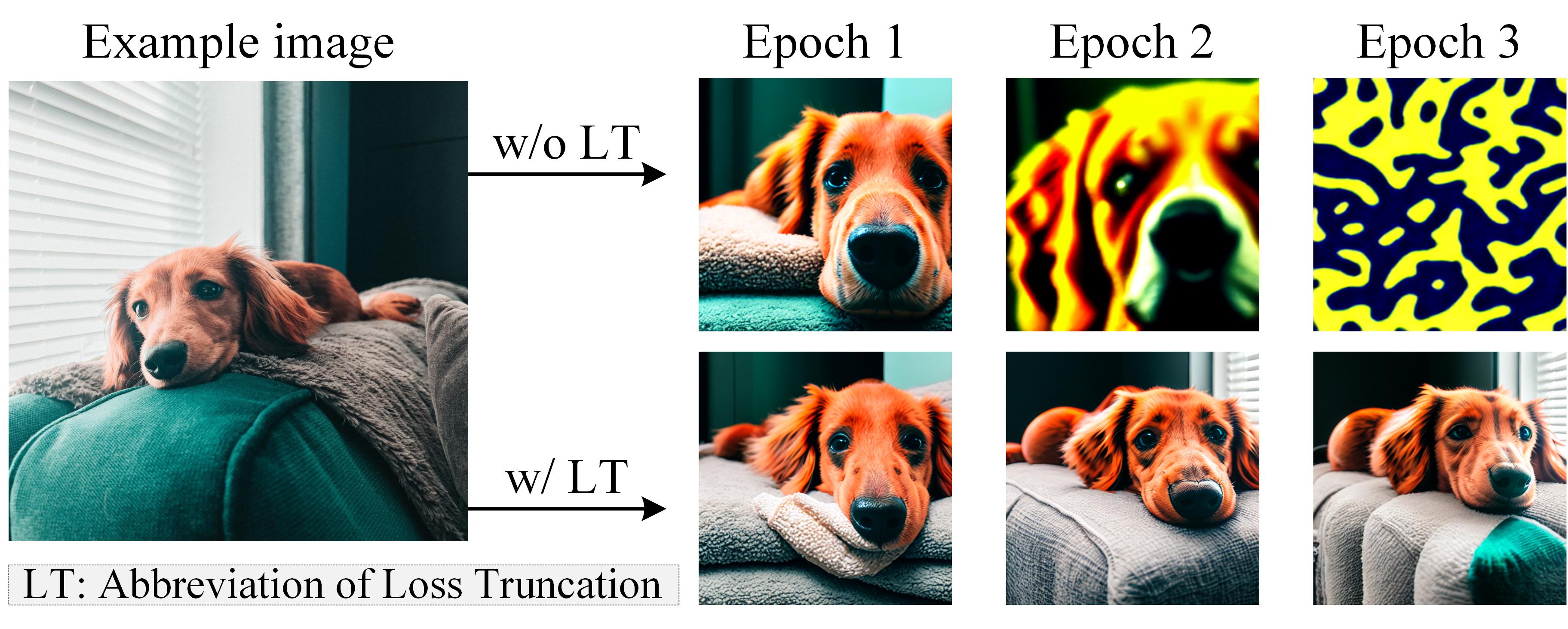}}
\vspace{-0.5em}
\caption{\textbf{Loss truncation.} SuDe-generations with and without truncation using Custom Diffusion as the baseline.
}
\label{fig: adaptive truncation}
\vspace{-1.0em}
\end{figure}

\textbf{Image fidelity:} 
Besides, our method still maintains subject fidelity while generating attribute-rich images.
For example, in the 1st row, the dog generated with SuDe is in a very different pose than the example image, but we still can be sure that they are the same dog due to their private attributes, e.g., the golden hair, facial features, etc.



\subsection{Quantitative Results}
Here we quantitatively verify the conclusion in Sec.~\ref{sec: Qualitative Results}.
As shown in Table~\ref{tab: Quantitative Results}, our SuDe achieves stable improvement on attribute alignment, i.e., BLIP-T under SD-v1.4 and SD-v1.5 of $4.2\%$ and $2.6\%$ on ViCo, $0.9\%$ and $2.0\%$ on Custom Diffusion, and $1.2\%$ and $1.5\%$ on Dreambooth.
Besides, we show the performances (marked by $\dagger$) of a flexible $w_s$ (best results from the [0.5, 1.0, 2.0] $\cdot$ $w_{s}$).
We see that this low-cost adjustment could further expand the improvements, i.e., BLIP-T under SD-v1.4 and SD-v1.5 of $5.3\%$ and $3.9\%$ on ViCo, $1.1\%$ and $2.3\%$ on Custom Diffusion, and $3.2\%$ and $2.0\%$ on Dreambooth.
More analysis about the $w_s$ is in Sec.~\ref{sec: ablation w}.
For the subject fidelity, SuDe only brings a slight fluctuation to the baseline's DINO-I, indicating that our method will not sacrifice the subject fidelity.

\subsection{Empirical Study}
\subsubsection{Training weight $w_{s}$} \label{sec: ablation w}
The $w_{s}$ affects the weight proportion of $\mathcal{L}_{sude}$.
We visualize the generated image under different $w_{s}$ in Fig.~\ref{fig: ablation w}, by which we can summarize that:
\textbf{1)} As the $w_s$ increases, the subject (e.g., teapot) can inherit public attributes (e.g., clear) more comprehensively.
A $w_s$ within an appropriate range (e.g., $[0.5, 2] \cdot w_s $ for the teapot) could preserve the subject fidelity well.
But a too-large $w_s$ causes our model to lose subject fidelity (e.g., 4 $\cdot w_s$ for the bowl) since it dilutes the $\mathcal{L}_{sub}$ for learning private attributes.
\textbf{2)} A small $w_s$ is more proper for an attribute-simple subject (e.g., bowl), while a large $w_s$ is more proper for an attribute-complex subject (e.g., dog).
Another interesting phenomenon in Fig.~\ref{fig: ablation w} 1st line is that the baseline generates images with berries, but our SuDe does not.
This is because though the berry appears in the example, it is not an attribute of the bowl, thus it is not captured by our derived class modeling.
Further, in Sec.~\ref{sec: attribute-unrelated}, we show that our method can also combine attribute-related and attribute-unrelated generations with the help of prompts, where one can make customizations like `photo of a metal \{$S*$\} with cherry'.

\begin{figure}[htbp]
\centerline{\includegraphics[scale=0.7]{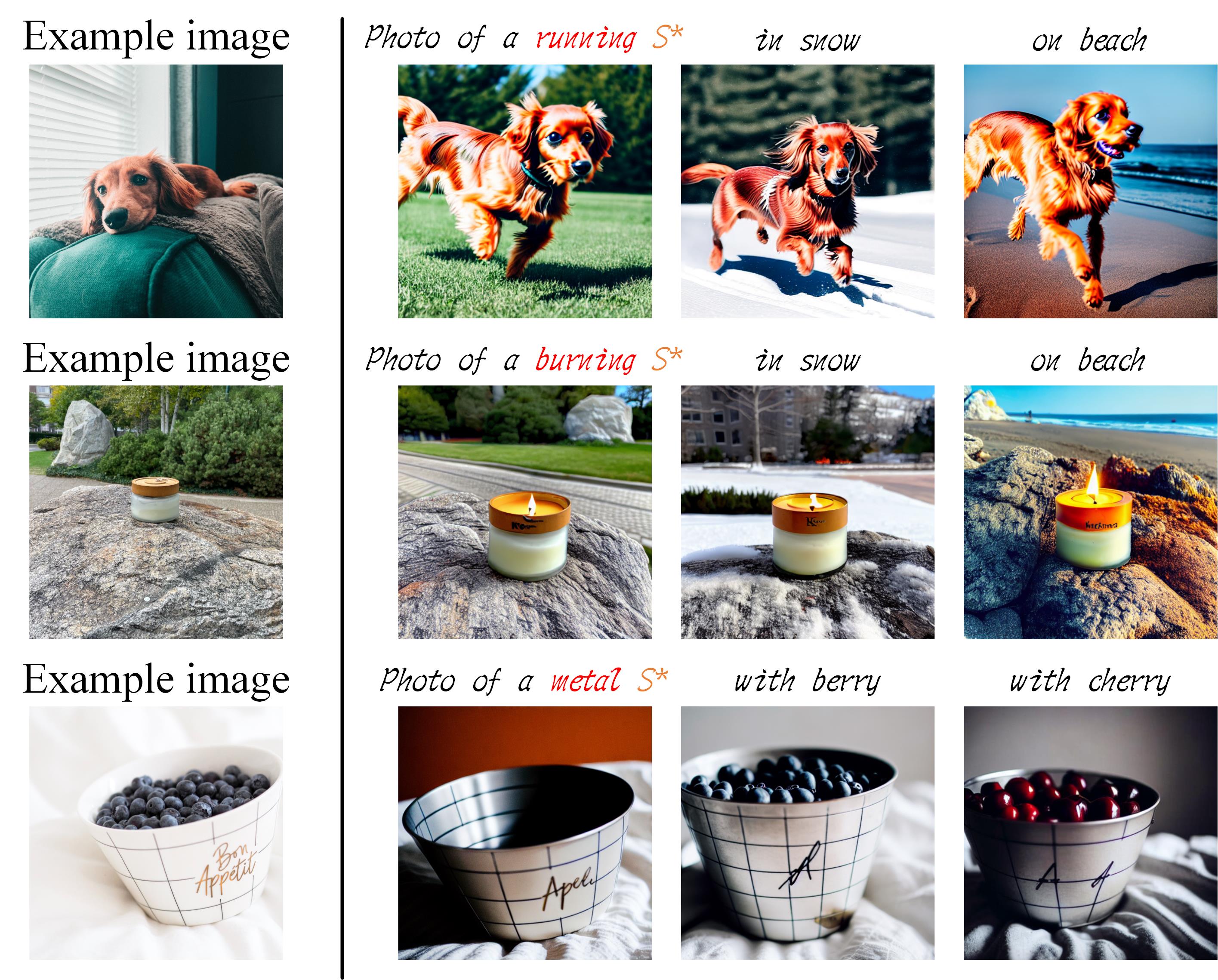}}
\vspace{-0.5em}
\caption{\textbf{Combine with attribute-unrelated prompts.} Generations with both attribute-related and attribute-unrelated prompts.
}
\label{fig: attribute_with_background}
\end{figure}

\begin{table}[thbp]
\setlength{\tabcolsep}{0.7mm}
  \centering
  \caption{\textbf{The BLIP-T computed with various prompt templates.} The $\bm{P}_{0}$ is the baseline's default prompt of `photo of a [attribute] \{S$^{*}$\}', and $\bm{P}_{1}$ to $\bm{P}_{3}$ are described in Sec.~\ref{sec: prompt}.
  }\label{tab: ablation prompt}
\begin{tabular}{l|cccc}
\hline
Prompt   & $\bm{P}_{0}$   & $\bm{P}_{1}$   & $\bm{P}_{2}$   & $\bm{P}_{3}$     \\ \hline
ViCo~\cite{hao2023vico} & 39.1 & 40.8 & 40.9 & 41.2  \\
w/ SuDe  & 43.3 (\textcolor{red}{+4.2}) & 43.4 (\textcolor{red}{+2.6}) & 43.1 (\textcolor{red}{+2.2}) & 42.7 (\textcolor{red}{+1.5}) \\ \hline
\end{tabular}
\vspace{-1.0em}
\end{table}

\vspace{-1.0em}
\subsubsection{Ablation of loss truncation} \label{sec: ablation trunction}
In Sec.\ref{sec: Loss Value Truncate}, the loss truncation is designed to prevent the $p(\bm{c}_{cate}|x_{\theta}(\bm{x}_{t}, \bm{c}_{sub}, t))$ from over-optimization.
Here we verify that this truncation is important for preventing the training from collapsing. 
As Fig.~\ref{fig: adaptive truncation} shows, without truncation, the generations exhibit distortion at epoch 2 and completely collapse at epoch 3.
This is because over-optimizing $p(\bm{c}_{cate}|x_{\theta}(\bm{x}_{t}, \bm{c}_{sub}, t))$ makes a noisy image have an exorbitant classification probability.
An extreme example is classifying a pure noise into a certain category with a probability of 1.
This damages the semantic space of the pre-trained diffusion model, leading to generation collapse.



\subsubsection{Combine with attribute-unrelated prompts} \label{sec: attribute-unrelated}
In the above sections, we mainly demonstrated the advantages of our SuDe for attribute-related generations.
Here we show that our approach's advantage can also be combined with attribute-unrelated prompts for more imaginative customizations. 
As shown in Fig.~\ref{fig: attribute_with_background}, our method can generate images harmoniously like, a \{$S^{*}$\} (dog) running in various backgrounds, a \{$S^{*}$\} (candle) burning in various backgrounds, and a \{$S^{*}$\} metal (bowl) with various fruits.

\begin{figure}[htbp]
\centerline{\includegraphics[scale=0.78]{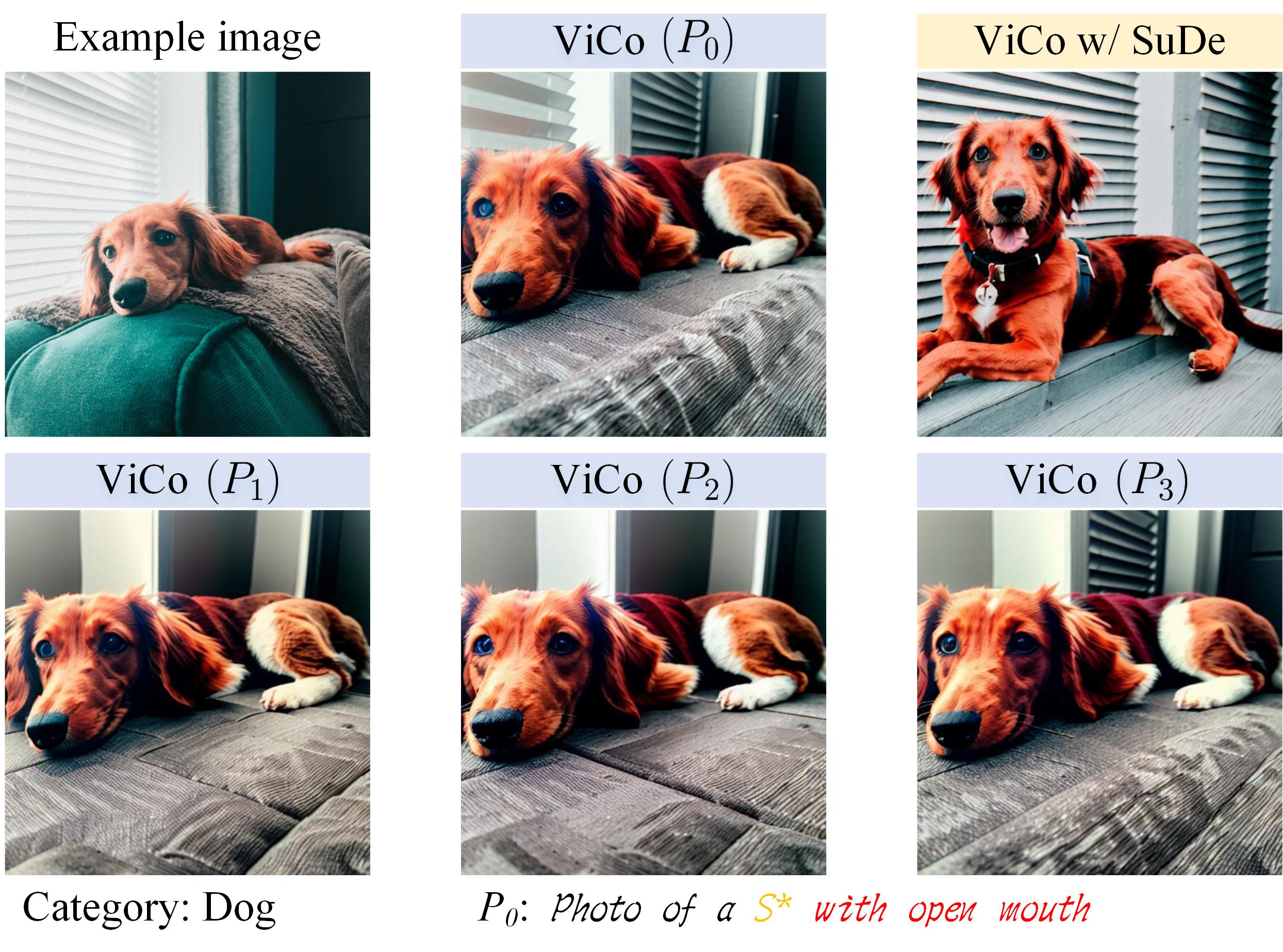}}
\vspace{-0.5em}
\caption{\textbf{Generations with various prompts.} The subject is a dog and the attribute we want to edit is `open mouth'.
$\bm{P}_0$ is the default prompt, and $\bm{P}_1$ to $\bm{P}_3$ are described in Sec.~\ref{sec: prompt}.
}
\label{fig: ablation prompt}
\vspace{-1.0em}
\end{figure}

\begin{figure}[htbp]
\centerline{\includegraphics[scale=0.94]{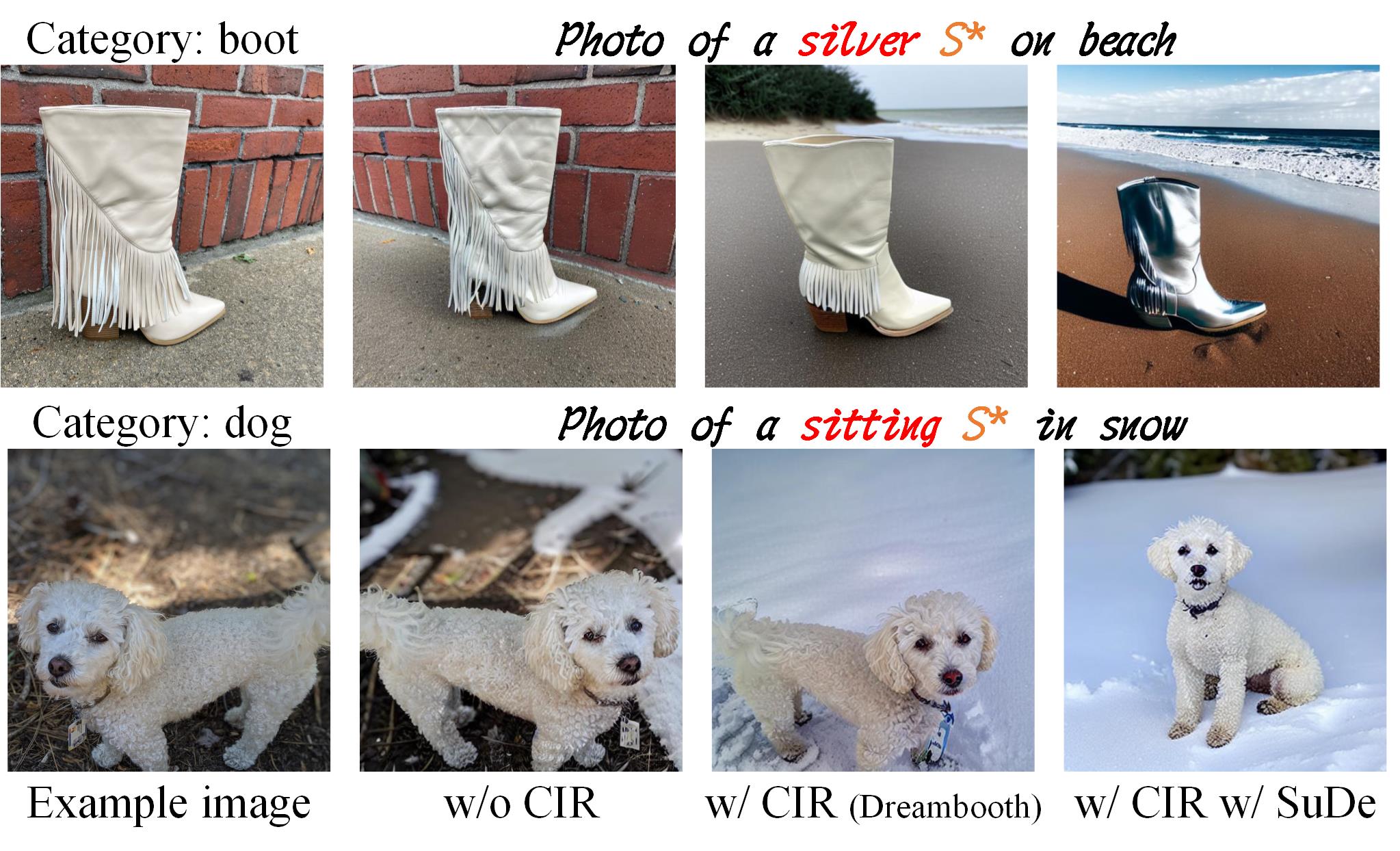}}
\vspace{-0.5em}
\caption{`CIR' is the abbreviation for class image regularization.
}
\label{fig: CIR}
\vspace{-1.0em}
\end{figure}





\subsubsection{Compare with class image regularization} \label{sec: CIR}
In existing subject-driven generation methods~\cite{ruiz2023dreambooth,hao2023vico,kumari2023multi}, as mentioned in Eq.~\ref{eq: total loss}, a regularization item $\mathcal{L}_{reg}$ is usually used to prevent the model overfitting to the subject example.
Here we discuss the difference between the roles of $\mathcal{L}_{reg}$ and our $\mathcal{L}_{sude}$.
Using the class image regularization $\mathcal{L}_{reg}$ in DreamBooth as an example, it is defined as:
\begin{equation}
  \begin{aligned}
  \mathcal{L}_{reg} = ||x_{\Bar{\theta}_{pr}}(\bm{x}_{t}, \bm{c}_{cate}, t) - x_{\theta}(\bm{x}_{t}, \bm{c}_{cate}, t)||^{2},
  \end{aligned} \label{eq: CIR}
\end{equation}
where the $x_{\Bar{\theta}_{pr}}$ is the frozen pre-trained diffusion model.
It can be seen that Eq.~\ref{eq: CIR} enforces the generation conditioned on $\bm{c}_{cate}$ to keep the same before and after subject-driven finetuning.
Visually, based on Fig.~\ref{fig: CIR}, we find that the $\mathcal{L}_{reg}$ mainly benefits background editing.
But it only uses the `category prompt' ($\bm{c}_{cate}$) alone, ignoring modeling the affiliation between $\bm{c}_{sub}$ and $\bm{c}_{cate}$.
Thus it cannot benefit attribute editing like our SuDe.


\vspace{-1.0em}

\subsubsection{Compare with modifying prompt} \label{sec: prompt}
Essentially, our SuDe enriches the concept of a subject by the public attributes of its category.
A naive alternative to realize this is to provide both the subject token and category token in the text prompt, e.g., `photo of a \{S$^{*}$\} [category]', which is already used in the DreamBooth~\cite{ruiz2023dreambooth} and Custom Diffusion~\cite{kumari2023multi} baselines.
The above comparisons on these two baselines show that this kind of prompt cannot tackle the attribute-missing problem well.
Here we further evaluate the performances of other prompt projects on the ViCo baseline, since its default prompt only contains the subject token.
Specifically, we verify three prompt templates: $\bm{P_1}$: `photo of a [attribute] \{S$^{*}$\} [category]', $\bm{P_2}$: `photo of a [attribute] \{S$^{*}$\} and it is a [category]', $\bm{P_3}$: `photo of a \{S$^{*}$\} and it is a [attribute] [category]'.
Referring to works in prompt learning~\cite{schick2021exploiting,liu2023pre,petroni2019language,song2023augprompt}, we retained the triggering word structure in these templates, the form of `photo of a \{S$^{*}$\}' that was used in subject-driven finetuning.


As shown in Table~\ref{tab: ablation prompt}, a good prompt template can partly alleviate this problem, e.g., $\bm{P_3}$ gets a BLIP-T of 41.2.
But there are still some attributes that cannot be supplied by modifying prompt, e.g., in Fig.~\ref{fig: ablation prompt}, $\bm{P_1}$ to $\bm{P_3}$ cannot make the dog with `open mouth'.
This is because they only put both subject and category in the prompt, but ignore modeling their relationships like our SuDe.
Besides, our method can also work on these prompt templates, as in Table~\ref{tab: ablation prompt}, SuDe further improves all prompts by over $1.5\%$.









\section{Conclusion}
In this paper, we creatively model subject-driven generation as building a derived class.
Specifically, we propose subject-derived regularization (SuDe) to make a subject inherit public attributes from its semantic category while learning its private attributes from the subject example.
As a plugin-and-play method, our SuDe can conveniently combined with existing baselines and improve attribute-related generations.
Our SuDe faces the most challenging but valuable one-shot scene and can generate imaginative customizations, showcasing attractive application prospects.

\textbf{Broader Impact.} 
Subject-driven generation is a newly emerging application, most works of which currently focus on image customizations with attribute-unrelated prompts.
But a foreseeable and valuable scenario is to make more modal customizations with the user-provided image, where attribute-related generation will be widely needed.
This paper proposes the modeling that builds a subject as a derived class of its semantic category, enabling good attribute-related generations, and thereby providing a promising solution for future subject-driven applications.

\textbf{Acknowledgments.}
We extend our gratitude to the FaceChain community for their contributions to this work.

{
    \small
    \bibliographystyle{ieeenat_fullname}
    \bibliography{main}

\begin{thebibliography}{44}
\providecommand{\natexlab}[1]{#1}
\providecommand{\url}[1]{\texttt{#1}}
\expandafter\ifx\csname urlstyle\endcsname\relax
  \providecommand{\doi}[1]{doi: #1}\else
  \providecommand{\doi}{doi: \begingroup \urlstyle{rm}\Url}\fi

\bibitem[Uns()]{Unsplash}
Unsplash.
\newblock In \emph{\url{https://unsplash.com/}}.

\bibitem[str(1988)]{stroustrup1988object}
What is object-oriented programming?
\newblock \emph{IEEE software}, 5\penalty0 (3):\penalty0 10--20, 1988.

\bibitem[202(2022)]{2022sd}
Stable diffusion.
\newblock In \emph{\url{https : / / huggingface . co / CompVis/stable-diffusion-v-1-4-original}}, 2022.

\bibitem[Balaji et~al.(2022)Balaji, Nah, Huang, Vahdat, Song, Zhang, Kreis, Aittala, Aila, Laine, et~al.]{balaji2022ediff}
Yogesh Balaji, Seungjun Nah, Xun Huang, Arash Vahdat, Jiaming Song, Qinsheng Zhang, Karsten Kreis, Miika Aittala, Timo Aila, Samuli Laine, et~al.
\newblock ediff-i: Text-to-image diffusion models with an ensemble of expert denoisers.
\newblock \emph{arXiv preprint arXiv:2211.01324}, 2022.

\bibitem[Caron et~al.(2021)Caron, Touvron, Misra, J{\'e}gou, Mairal, Bojanowski, and Joulin]{caron2021emerging}
Mathilde Caron, Hugo Touvron, Ishan Misra, Herv{\'e} J{\'e}gou, Julien Mairal, Piotr Bojanowski, and Armand Joulin.
\newblock Emerging properties in self-supervised vision transformers.
\newblock In \emph{Int. Conf. Comput. Vis.}, pages 9650--9660, 2021.

\bibitem[Chang et~al.(2023)Chang, Zhang, Barber, Maschinot, Lezama, Jiang, Yang, Murphy, Freeman, Rubinstein, et~al.]{chang2023muse}
Huiwen Chang, Han Zhang, Jarred Barber, AJ Maschinot, Jose Lezama, Lu Jiang, Ming-Hsuan Yang, Kevin Murphy, William~T Freeman, Michael Rubinstein, et~al.
\newblock Muse: Text-to-image generation via masked generative transformers.
\newblock \emph{arXiv preprint arXiv:2301.00704}, 2023.

\bibitem[Chen et~al.(2023{\natexlab{a}})Chen, Zhang, Wang, Duan, Zhou, and Zhu]{chen2023disenbooth}
Hong Chen, Yipeng Zhang, Xin Wang, Xuguang Duan, Yuwei Zhou, and Wenwu Zhu.
\newblock Disenbooth: Disentangled parameter-efficient tuning for subject-driven text-to-image generation.
\newblock \emph{arXiv preprint arXiv:2305.03374}, 2023{\natexlab{a}}.

\bibitem[Chen et~al.(2023{\natexlab{b}})Chen, Hu, Li, Rui, Jia, Chang, and Cohen]{chen2023subject}
Wenhu Chen, Hexiang Hu, Yandong Li, Nataniel Rui, Xuhui Jia, Ming-Wei Chang, and William~W Cohen.
\newblock Subject-driven text-to-image generation via apprenticeship learning.
\newblock \emph{arXiv preprint arXiv:2304.00186}, 2023{\natexlab{b}}.

\bibitem[Crowson et~al.(2022)Crowson, Biderman, Kornis, Stander, Hallahan, Castricato, and Raff]{crowson2022vqgan}
Katherine Crowson, Stella Biderman, Daniel Kornis, Dashiell Stander, Eric Hallahan, Louis Castricato, and Edward Raff.
\newblock Vqgan-clip: Open domain image generation and editing with natural language guidance.
\newblock In \emph{Eur. Conf. Comput. Vis.}, pages 88--105. Springer, 2022.

\bibitem[Ding et~al.(2021)Ding, Yang, Hong, Zheng, Zhou, Yin, Lin, Zou, Shao, Yang, et~al.]{ding2021cogview}
Ming Ding, Zhuoyi Yang, Wenyi Hong, Wendi Zheng, Chang Zhou, Da Yin, Junyang Lin, Xu Zou, Zhou Shao, Hongxia Yang, et~al.
\newblock Cogview: Mastering text-to-image generation via transformers.
\newblock \emph{Adv. Neural Inform. Process. Syst.}, 34:\penalty0 19822--19835, 2021.

\bibitem[Ding et~al.(2022)Ding, Zheng, Hong, and Tang]{ding2022cogview2}
Ming Ding, Wendi Zheng, Wenyi Hong, and Jie Tang.
\newblock Cogview2: Faster and better text-to-image generation via hierarchical transformers.
\newblock \emph{Adv. Neural Inform. Process. Syst.}, 35:\penalty0 16890--16902, 2022.

\bibitem[Gafni et~al.(2022)Gafni, Polyak, Ashual, Sheynin, Parikh, and Taigman]{gafni2022make}
Oran Gafni, Adam Polyak, Oron Ashual, Shelly Sheynin, Devi Parikh, and Yaniv Taigman.
\newblock Make-a-scene: Scene-based text-to-image generation with human priors.
\newblock In \emph{Eur. Conf. Comput. Vis.}, pages 89--106. Springer, 2022.

\bibitem[Gal et~al.(2022)Gal, Alaluf, Atzmon, Patashnik, Bermano, Chechik, and Cohen-or]{gal2022image}
Rinon Gal, Yuval Alaluf, Yuval Atzmon, Or Patashnik, Amit~Haim Bermano, Gal Chechik, and Daniel Cohen-or.
\newblock An image is worth one word: Personalizing text-to-image generation using textual inversion.
\newblock In \emph{Int. Conf. Learn. Represent.}, 2022.

\bibitem[Hao et~al.(2023)Hao, Han, Zhao, and Wong]{hao2023vico}
Shaozhe Hao, Kai Han, Shihao Zhao, and Kwan-Yee~K Wong.
\newblock Vico: Detail-preserving visual condition for personalized text-to-image generation.
\newblock \emph{arXiv preprint arXiv:2306.00971}, 2023.

\bibitem[Ho et~al.(2020)Ho, Jain, and Abbeel]{ho2020denoising}
Jonathan Ho, Ajay Jain, and Pieter Abbeel.
\newblock Denoising diffusion probabilistic models.
\newblock \emph{Adv. Neural Inform. Process. Syst.}, 33:\penalty0 6840--6851, 2020.

\bibitem[JOYCE(2003)]{joyce2003bayes}
J JOYCE.
\newblock Bayes' theorem.
\newblock \emph{Stanford Encyclopedia of Philosophy}, 2003.

\bibitem[Kenton and Toutanova(2019)]{kenton2019bert}
Jacob Devlin Ming-Wei~Chang Kenton and Lee~Kristina Toutanova.
\newblock Bert: Pre-training of deep bidirectional transformers for language understanding.
\newblock In \emph{Proceedings of NAACL-HLT}, pages 4171--4186, 2019.

\bibitem[Kumari et~al.(2023)Kumari, Zhang, Zhang, Shechtman, and Zhu]{kumari2023multi}
Nupur Kumari, Bingliang Zhang, Richard Zhang, Eli Shechtman, and Jun-Yan Zhu.
\newblock Multi-concept customization of text-to-image diffusion.
\newblock In \emph{IEEE Conf. Comput. Vis. Pattern Recog.}, pages 1931--1941, 2023.

\bibitem[Li et~al.(2022)Li, Li, Xiong, and Hoi]{li2022blip}
Junnan Li, Dongxu Li, Caiming Xiong, and Steven Hoi.
\newblock Blip: Bootstrapping language-image pre-training for unified vision-language understanding and generation.
\newblock In \emph{International Conference on Machine Learning}, pages 12888--12900. PMLR, 2022.

\bibitem[Liu et~al.(2023{\natexlab{a}})Liu, Yuan, Fu, Jiang, Hayashi, and Neubig]{liu2023pre}
Pengfei Liu, Weizhe Yuan, Jinlan Fu, Zhengbao Jiang, Hiroaki Hayashi, and Graham Neubig.
\newblock Pre-train, prompt, and predict: A systematic survey of prompting methods in natural language processing.
\newblock \emph{ACM Computing Surveys}, 55\penalty0 (9):\penalty0 1--35, 2023{\natexlab{a}}.

\bibitem[Liu et~al.(2023{\natexlab{b}})Liu, Zeng, Ren, Li, Zhang, Yang, Li, Yang, Su, Zhu, et~al.]{liu2023grounding}
Shilong Liu, Zhaoyang Zeng, Tianhe Ren, Feng Li, Hao Zhang, Jie Yang, Chunyuan Li, Jianwei Yang, Hang Su, Jun Zhu, et~al.
\newblock Grounding dino: Marrying dino with grounded pre-training for open-set object detection.
\newblock \emph{arXiv preprint arXiv:2303.05499}, 2023{\natexlab{b}}.

\bibitem[Nichol et~al.(2022)Nichol, Dhariwal, Ramesh, Shyam, Mishkin, Mcgrew, Sutskever, and Chen]{nichol2022glide}
Alexander~Quinn Nichol, Prafulla Dhariwal, Aditya Ramesh, Pranav Shyam, Pamela Mishkin, Bob Mcgrew, Ilya Sutskever, and Mark Chen.
\newblock Glide: Towards photorealistic image generation and editing with text-guided diffusion models.
\newblock In \emph{International Conference on Machine Learning}, pages 16784--16804. PMLR, 2022.

\bibitem[Petroni et~al.(2019)Petroni, Rockt{\"a}schel, Lewis, Bakhtin, Wu, Miller, and Riedel]{petroni2019language}
F Petroni, T Rockt{\"a}schel, P Lewis, A Bakhtin, Y Wu, AH Miller, and S Riedel.
\newblock Language models as knowledge bases?
\newblock Association for Computational Linguistics, 2019.

\bibitem[Radford et~al.(2021)Radford, Kim, Hallacy, Ramesh, Goh, Agarwal, Sastry, Askell, Mishkin, Clark, et~al.]{radford2021learning}
Alec Radford, Jong~Wook Kim, Chris Hallacy, Aditya Ramesh, Gabriel Goh, Sandhini Agarwal, Girish Sastry, Amanda Askell, Pamela Mishkin, Jack Clark, et~al.
\newblock Learning transferable visual models from natural language supervision.
\newblock In \emph{International Conference on Machine Learning}, pages 8748--8763. PMLR, 2021.

\bibitem[Ramesh et~al.(2021)Ramesh, Pavlov, Goh, Gray, Voss, Radford, Chen, and Sutskever]{ramesh2021zero}
Aditya Ramesh, Mikhail Pavlov, Gabriel Goh, Scott Gray, Chelsea Voss, Alec Radford, Mark Chen, and Ilya Sutskever.
\newblock Zero-shot text-to-image generation.
\newblock In \emph{International Conference on Machine Learning}, pages 8821--8831. PMLR, 2021.

\bibitem[Ramesh et~al.(2022)Ramesh, Dhariwal, Nichol, Chu, and Chen]{ramesh2022hierarchical}
Aditya Ramesh, Prafulla Dhariwal, Alex Nichol, Casey Chu, and Mark Chen.
\newblock Hierarchical text-conditional image generation with clip latents.
\newblock \emph{arXiv preprint arXiv:2204.06125}, 2022.

\bibitem[Reed et~al.(2016)Reed, Akata, Yan, Logeswaran, Schiele, and Lee]{reed2016generative}
Scott Reed, Zeynep Akata, Xinchen Yan, Lajanugen Logeswaran, Bernt Schiele, and Honglak Lee.
\newblock Generative adversarial text to image synthesis.
\newblock In \emph{International Conference on Machine Learning}, pages 1060--1069. PMLR, 2016.

\bibitem[Rentsch(1982)]{rentsch1982object}
Tim Rentsch.
\newblock Object oriented programming.
\newblock \emph{ACM Sigplan Notices}, 17\penalty0 (9):\penalty0 51--57, 1982.

\bibitem[Rombach et~al.(2022)Rombach, Blattmann, Lorenz, Esser, and Ommer]{rombach2022high}
Robin Rombach, Andreas Blattmann, Dominik Lorenz, Patrick Esser, and Bj{\"o}rn Ommer.
\newblock High-resolution image synthesis with latent diffusion models.
\newblock In \emph{IEEE Conf. Comput. Vis. Pattern Recog.}, pages 10684--10695, 2022.

\bibitem[Ruiz et~al.(2023{\natexlab{a}})Ruiz, Li, Jampani, Pritch, Rubinstein, and Aberman]{ruiz2023dreambooth}
Nataniel Ruiz, Yuanzhen Li, Varun Jampani, Yael Pritch, Michael Rubinstein, and Kfir Aberman.
\newblock Dreambooth: Fine tuning text-to-image diffusion models for subject-driven generation.
\newblock In \emph{IEEE Conf. Comput. Vis. Pattern Recog.}, pages 22500--22510, 2023{\natexlab{a}}.

\bibitem[Ruiz et~al.(2023{\natexlab{b}})Ruiz, Li, Jampani, Wei, Hou, Pritch, Wadhwa, Rubinstein, and Aberman]{ruiz2023hyperdreambooth}
Nataniel Ruiz, Yuanzhen Li, Varun Jampani, Wei Wei, Tingbo Hou, Yael Pritch, Neal Wadhwa, Michael Rubinstein, and Kfir Aberman.
\newblock Hyperdreambooth: Hypernetworks for fast personalization of text-to-image models.
\newblock \emph{arXiv preprint arXiv:2307.06949}, 2023{\natexlab{b}}.

\bibitem[Saharia et~al.(2022)Saharia, Chan, Saxena, Li, Whang, Denton, Ghasemipour, Gontijo~Lopes, Karagol~Ayan, Salimans, et~al.]{saharia2022photorealistic}
Chitwan Saharia, William Chan, Saurabh Saxena, Lala Li, Jay Whang, Emily~L Denton, Kamyar Ghasemipour, Raphael Gontijo~Lopes, Burcu Karagol~Ayan, Tim Salimans, et~al.
\newblock Photorealistic text-to-image diffusion models with deep language understanding.
\newblock \emph{Adv. Neural Inform. Process. Syst.}, 35:\penalty0 36479--36494, 2022.

\bibitem[Schick and Sch{\"u}tze(2021)]{schick2021exploiting}
Timo Schick and Hinrich Sch{\"u}tze.
\newblock Exploiting cloze-questions for few-shot text classification and natural language inference.
\newblock In \emph{Proceedings of the Conference of the European Chapter of the Association for Computational Linguistics}, pages 255--269, 2021.

\bibitem[Sohl-Dickstein et~al.(2015)Sohl-Dickstein, Weiss, Maheswaranathan, and Ganguli]{sohl2015deep}
Jascha Sohl-Dickstein, Eric Weiss, Niru Maheswaranathan, and Surya Ganguli.
\newblock Deep unsupervised learning using nonequilibrium thermodynamics.
\newblock In \emph{International Conference on Machine Learning}, pages 2256--2265. PMLR, 2015.

\bibitem[Song et~al.(2023)Song, Cai, Zheng, Zhao, and Shao]{song2023augprompt}
Chengyu Song, Fei Cai, Jianming Zheng, Xiang Zhao, and Taihua Shao.
\newblock Augprompt: Knowledgeable augmented-trigger prompt for few-shot event classification.
\newblock \emph{Information Processing \& Management}, 60\penalty0 (4):\penalty0 103153, 2023.

\bibitem[Song et~al.(2020)Song, Meng, and Ermon]{song2020denoising}
Jiaming Song, Chenlin Meng, and Stefano Ermon.
\newblock Denoising diffusion implicit models.
\newblock In \emph{Int. Conf. Learn. Represent.}, 2020.

\bibitem[Stroustrup(1986)]{stroustrup1986overview}
Bjarne Stroustrup.
\newblock An overview of c++.
\newblock In \emph{Proceedings of the 1986 SIGPLAN workshop on Object-oriented programming}, pages 7--18, 1986.

\bibitem[Tao et~al.(2022)Tao, Tang, Wu, Jing, Bao, and Xu]{tao2022df}
Ming Tao, Hao Tang, Fei Wu, Xiao-Yuan Jing, Bing-Kun Bao, and Changsheng Xu.
\newblock Df-gan: A simple and effective baseline for text-to-image synthesis.
\newblock In \emph{IEEE Conf. Comput. Vis. Pattern Recog.}, pages 16515--16525, 2022.

\bibitem[Tewel et~al.(2023)Tewel, Gal, Chechik, and Atzmon]{tewel2023key}
Yoad Tewel, Rinon Gal, Gal Chechik, and Yuval Atzmon.
\newblock Key-locked rank one editing for text-to-image personalization.
\newblock In \emph{ACM SIGGRAPH 2023 Conference Proceedings}, pages 1--11, 2023.

\bibitem[Wegner(1990)]{wegner1990concepts}
Peter Wegner.
\newblock Concepts and paradigms of object-oriented programming.
\newblock \emph{ACM Sigplan Oops Messenger}, 1\penalty0 (1):\penalty0 7--87, 1990.

\bibitem[Wei et~al.(2023)Wei, Zhang, Ji, Bai, Zhang, and Zuo]{wei2023elite}
Yuxiang Wei, Yabo Zhang, Zhilong Ji, Jinfeng Bai, Lei Zhang, and Wangmeng Zuo.
\newblock Elite: Encoding visual concepts into textual embeddings for customized text-to-image generation.
\newblock 2023.

\bibitem[Xu et~al.(2018)Xu, Zhang, Huang, Zhang, Gan, Huang, and He]{xu2018attngan}
Tao Xu, Pengchuan Zhang, Qiuyuan Huang, Han Zhang, Zhe Gan, Xiaolei Huang, and Xiaodong He.
\newblock Attngan: Fine-grained text to image generation with attentional generative adversarial networks.
\newblock In \emph{IEEE Conf. Comput. Vis. Pattern Recog.}, pages 1316--1324, 2018.

\bibitem[Zhang et~al.(2023)Zhang, Dong, Tang, Huang, Huang, Ma, Lee, Deussen, and Xu]{zhang2023prospect}
Yuxin Zhang, Weiming Dong, Fan Tang, Nisha Huang, Haibin Huang, Chongyang Ma, Tong-Yee Lee, Oliver Deussen, and Changsheng Xu.
\newblock Prospect: Expanded conditioning for the personalization of attribute-aware image generation.
\newblock \emph{arXiv preprint arXiv:2305.16225}, 2023.

\bibitem[Zhu et~al.(2019)Zhu, Pan, Chen, and Yang]{zhu2019dm}
Minfeng Zhu, Pingbo Pan, Wei Chen, and Yi Yang.
\newblock Dm-gan: Dynamic memory generative adversarial networks for text-to-image synthesis.
\newblock In \emph{IEEE Conf. Comput. Vis. Pattern Recog.}, pages 5802--5810, 2019.

\end{thebibliography}
}

\clearpage
\maketitlesupplementary


\section{Overview}
We provide the dataset details in Sec.~\ref{sec: appendix dataset}.
Besides, we discuss the limitation of our SuDe in Sec.~\ref{sec: appendix limitation}.
For more empirical results, the details about the baselines' generations are in Sec.~\ref{sec: appendix base failure}, comparisons with offline method are in Sec.~\ref{sec: offline method}, more qualitative examples in Sec.~\ref{sec: appendix more example}, and the visualizations on more applications are in Sec.~\ref{sec: appendix more vis}.


\begin{figure}[thbp]
\centerline{\includegraphics[scale=0.5]{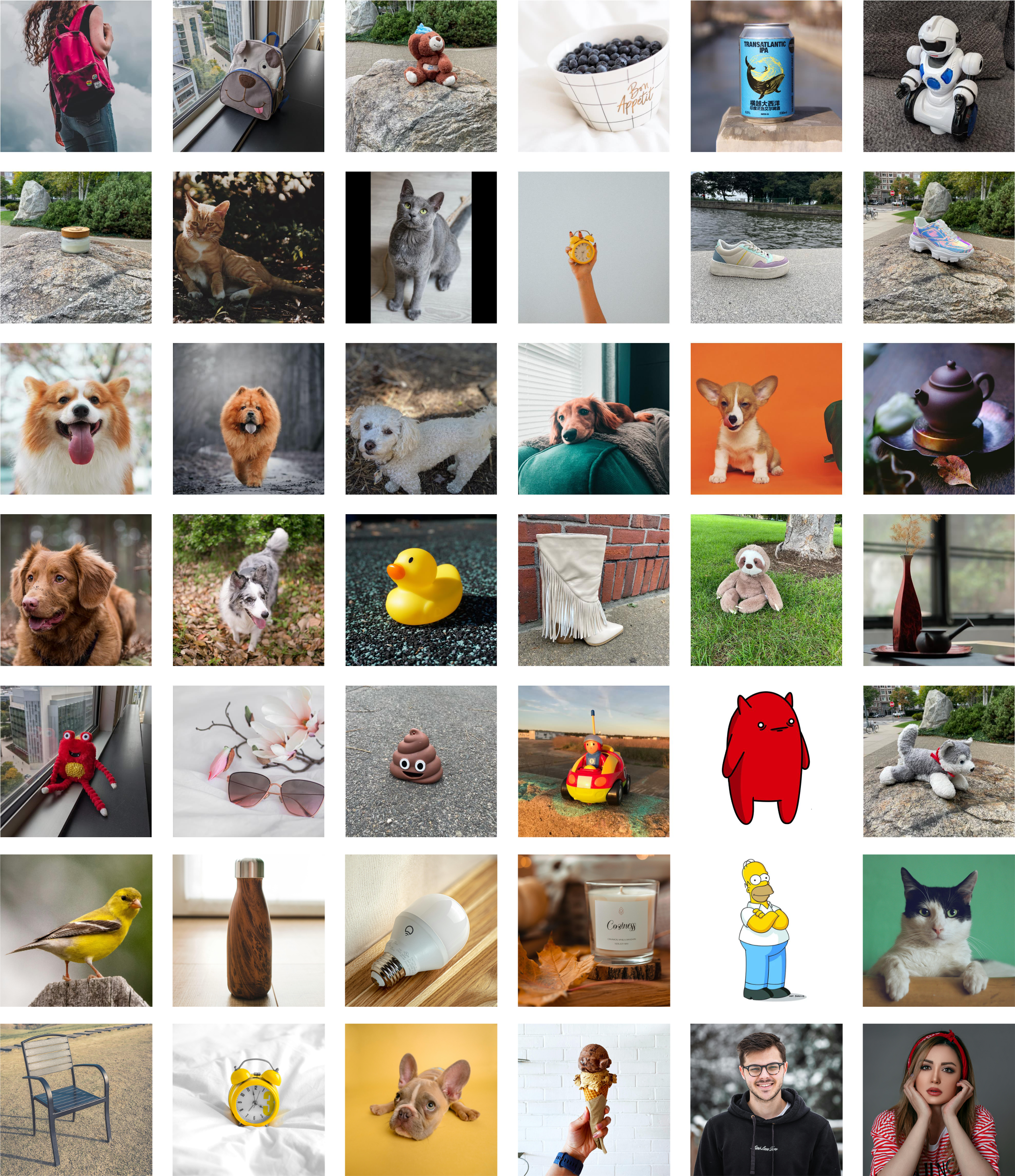}}
\caption{Subject image examples.}
\label{fig: dataset image}
\vspace{-1.0em}
\end{figure}

\section{Dataset Details} \label{sec: appendix dataset}
\subsection{Subject images}
For the images from the DreamBench~\cite{ruiz2023dreambooth}, which contains 30 subjects and 5 images for each subject, we only use one image (numbered '00.jpg') for each subject in all our experiments.
All the used images are shown in Fig.~\ref{fig: dataset image}.

\subsection{Prompts}
We collect 5 attribute-related prompts for all the 30 subjects.
The used prompts are shown in Table~\ref{tab: dataset prompt}.

\section{Limitation} \label{sec: appendix limitation}


\subsection{Inherent failure cases}
\begin{figure}[htbp]
\centerline{\includegraphics[scale=0.64]{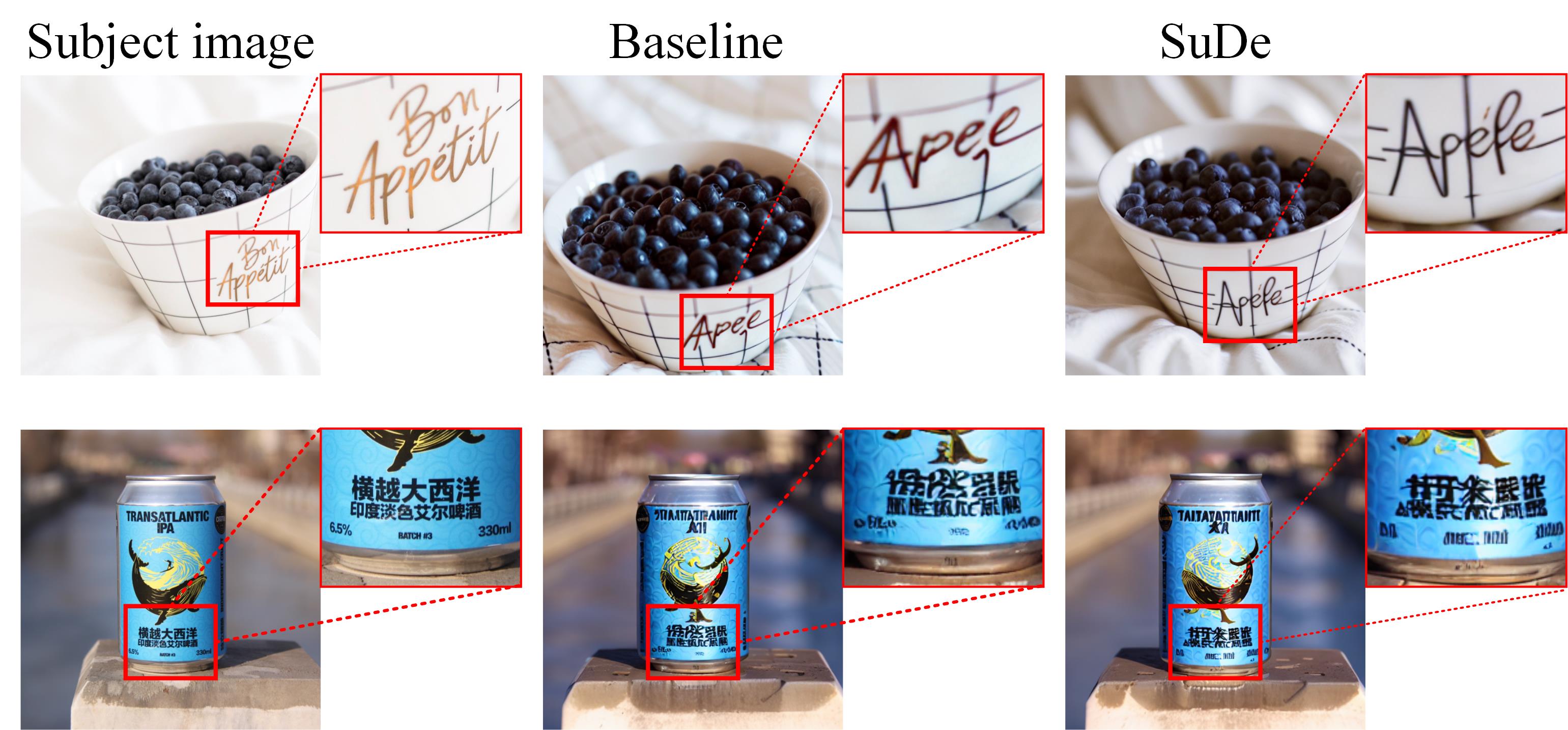}}
\caption{\textbf{Reconstruction results of texts.} The baseline here is Dreambooth~\cite{ruiz2023dreambooth}, and the prompt is `photo of a $S^*$'.}
\label{fig: inherent failure}
\vspace{-1.0em}
\end{figure}

As in Fig.~\ref{fig: inherent failure}, the text characters on the subject cannot be kept well, for both baselines w/ and w/o SuDe.
This is an inherent failure of the stable-diffusion backbone.
Our SuDe is designed to inherit the capabilities of the pre-trained model itself and therefore also inherits its shortcomings.

\subsection{Failure cases indirectly related to attributes}
\begin{figure}[hbp]
\vspace{-0.5em}
\centerline{\includegraphics[scale=0.54]{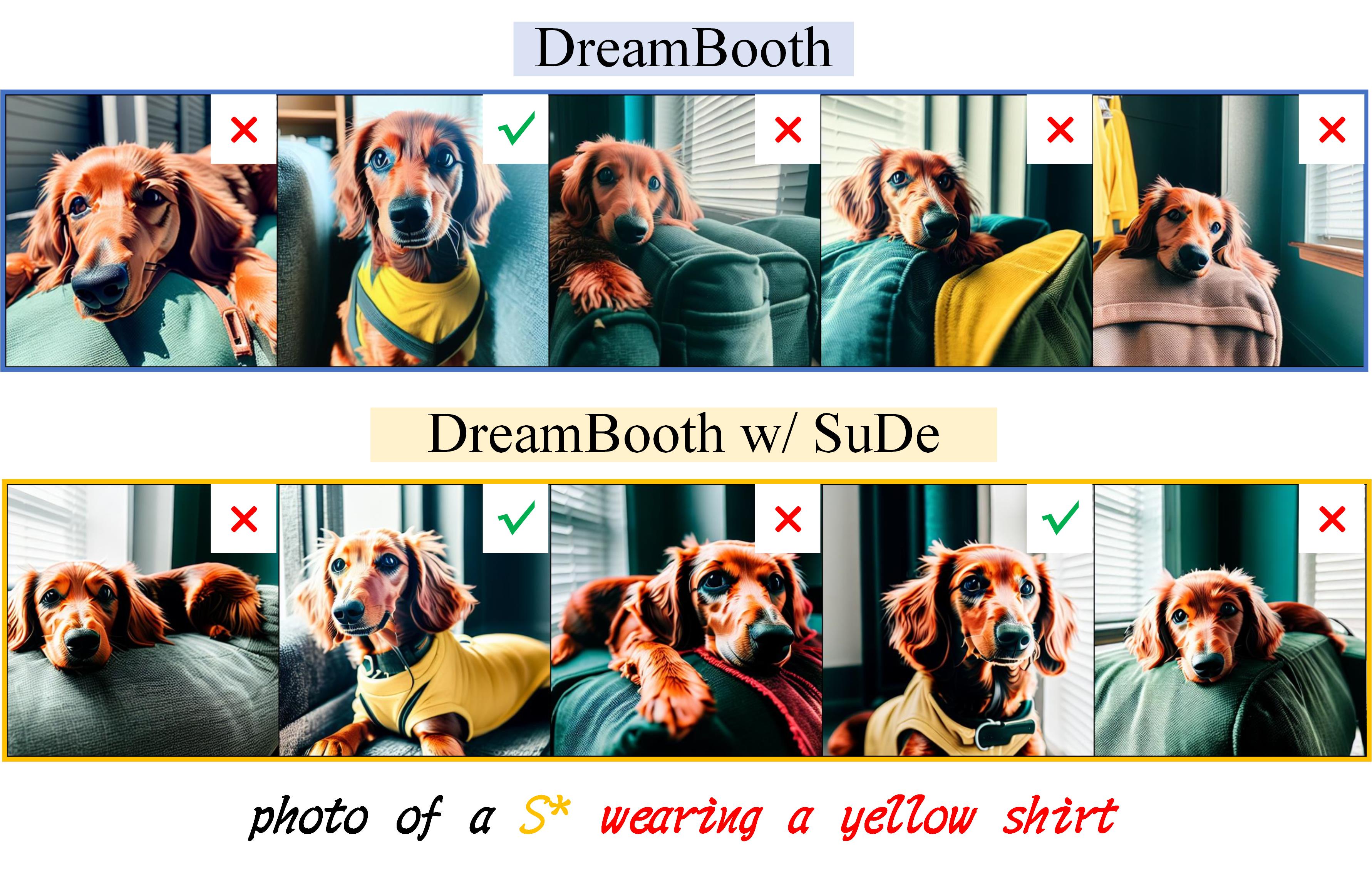}}
\caption{The 5 images are generated with various initial noises.}
\label{fig: appendix_limit_wearing}
\vspace{-1.0em}
\end{figure}

As Fig.~\ref{fig: appendix_limit_wearing}, the baseline model can only generate prompt-matching images with a very low probability (1 out of 5) for the prompt of `wearing a yellow shirt'.
For our SuDe, it performs better but is also not satisfactory enough.
This is because `wearing a shirt' is not a direct attribute of a dog, but is indirectly related to both the dog and the cloth.
Hence it cannot be directly inherited from the category attributes, thus our SuDe cannot solve this problem particularly well.

\begin{table*}[thbp]\footnotesize
\setlength{\tabcolsep}{0.3mm}
  \centering
  \caption{Prompts for each subject.
  }\label{tab: dataset prompt}
\begin{tabular}{cccccc}
\hline
\multicolumn{1}{l|}{Class}    & Backpack               & Stuffed animal              & Bowl                   & Can                         & Candle                                 \\ \hline
\multicolumn{1}{l|}{Prompt 1} & `photo of a blue \{\}'   & `photo of a blue \{\}'        & `photo of a blue \{\}'   & `photo of a blue \{\}'        & `photo of a burning \{\}'                \\
\multicolumn{1}{l|}{Prompt 2} & `photo of a green \{\}'  & `photo of a green \{\}'       & `photo of a green \{\}'  & `photo of a green \{\}'       & `photo of a cube shaped unburned \{\}'   \\
\multicolumn{1}{l|}{Prompt 3} & `photo of a yellow \{\}' & `photo of a yellow \{\}'      & `photo of a metal \{\}'  & `photo of a yellow \{\}'      & `photo of a cube shaped burning \{\}'    \\
\multicolumn{1}{l|}{Prompt 4} & `photo of a fallen \{\}' & `photo of a fallen \{\}'      & `photo of a shiny \{\}'  & `photo of a shiny \{\}'       & `photo of a burning \{\} with blue fire' \\
\multicolumn{1}{l|}{Prompt 5} & `photo of a dirty \{\}'  & `photo of a wet \{\}'         & `photo of a clear \{\}'  & `photo of a fallen \{\}'      & `photo of a blue\{\}'                    \\ \hline
\multicolumn{2}{c}{Cat}                                & Clock                       & Sneaker                & Toy                         & Dog                                    \\ \hline
\multicolumn{2}{c}{`photo of a running \{\}'}            & `photo of a blue \{\}'        & `photo of a blue \{\}'   & `photo of a blue \{\}'        & `photo of a running \{\}'                \\
\multicolumn{2}{c}{`photo of a jumping \{\}'}            & `photo of a green \{\}'       & `photo of a green \{\}'  & `photo of a green \{\}'       & `photo of a jumping \{\}'                \\
\multicolumn{2}{c}{`photo of a yawning \{\}'}            & `photo of a yellow \{\}'      & `photo of a yellow \{\}' & `photo of a yellow \{\}'      & `photo of a crawling \{\}'               \\
\multicolumn{2}{c}{`photo of a crawling \{\}'}           & `photo of a shiny \{\}'       & `photo of a red \{\}'    & `photo of a shiny \{\}'       & `photo of a \{\} with open mouth'        \\
\multicolumn{2}{c}{`photo of a \{\} climbing a tree'}    & `photo of a fallen \{\}'      & `photo of a white \{\}'  & `photo of a wet \{\}'         & `photo of a \{\} playing with a ball'    \\ \hline
\multicolumn{2}{c}{Teapot}                             & Glasses                     & Boot                   & Vase                        & Cartoon character                      \\ \hline
\multicolumn{2}{c}{`photo of a blue \{\}'}               & `photo of a blue \{\}'        & `photo of a blue \{\}'   & `photo of a blue \{\}'        & `photo of a running \{\}'                \\
\multicolumn{2}{c}{`photo of a shiny \{\}'}              & `photo of a green \{\}'       & `photo of a green \{\}'  & `photo of a green \{\}'       & `photo of a jumping \{\}'                \\
\multicolumn{2}{c}{`photo of a clear \{\}'}              & `photo of a yellow \{\}'      & `photo of a yellow \{\}' & `photo of a shiny \{\}'       & `photo of a \{\} swimming in pool'       \\
\multicolumn{2}{c}{`photo of a cube shaped \{\}'}        & `photo of a red \{\}'         & `photo of a shiny \{\}'  & `photo of a clear \{\}'       & `photo of a \{\} sleeping in bed'        \\
\multicolumn{2}{c}{`photo of a pumpkin shaped \{\}'}     & `photo of a cube shaped \{\}' & `photo of a wet \{\}'    & `photo of a cube shaped \{\}' & `photo of a \{\} driving a car'          \\ \hline
\end{tabular}
\end{table*}

\begin{figure*}[htbp]
\centerline{\includegraphics[scale=0.55]{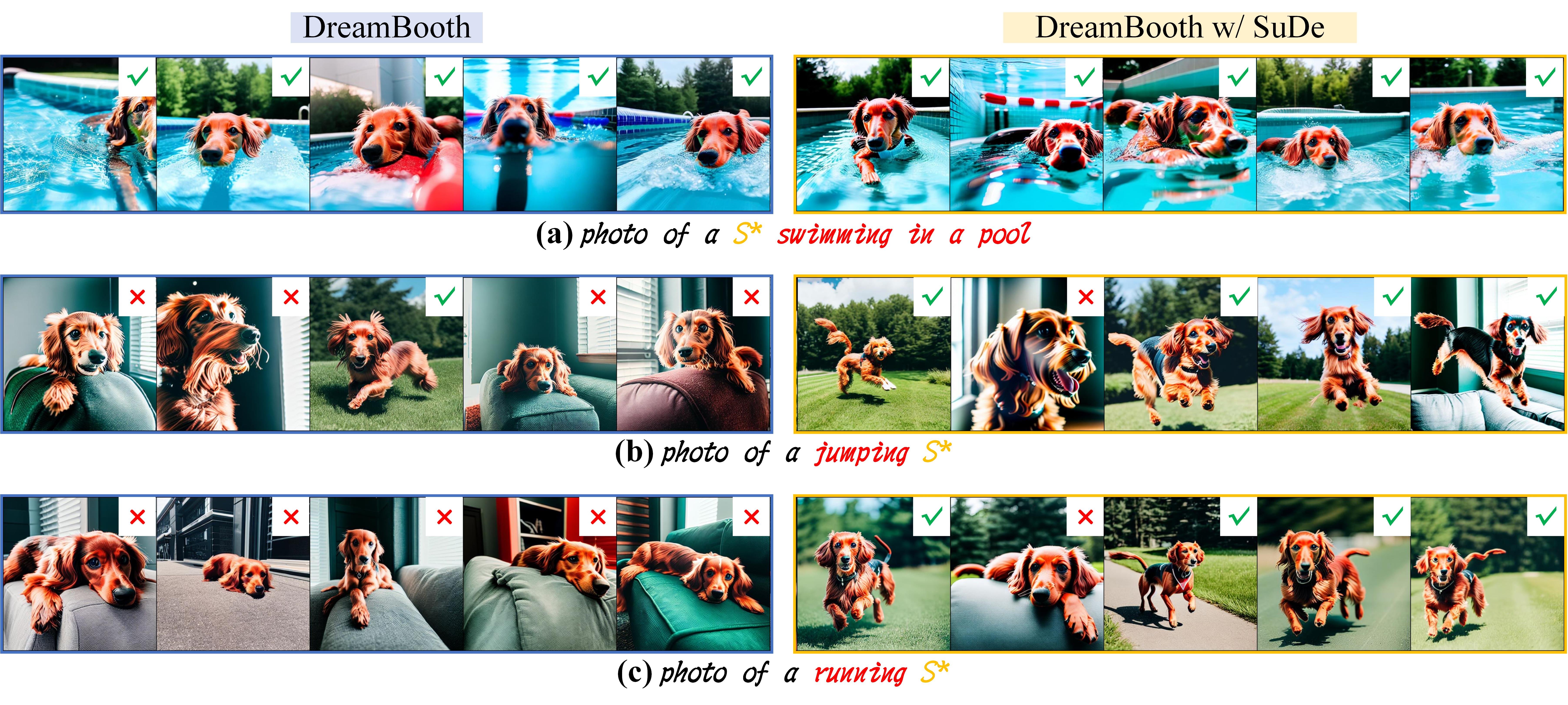}}
\vspace{-0.5em}
\caption{The subject image here is the dog shown in Fig.~\ref{fig: dataset image} line 3 and column 4. These results are generated by various initial noises.
}
\label{fig: appendix_base_cannot}
\vspace{-1.0em}
\end{figure*}

\section{More Experimental Results}
\subsection{Details about the generations of baselines} \label{sec: appendix base failure}
In the figures of the main manuscript, we mainly demonstrate the failure cases of the baseline, and our SuDe improves these cases.
In practice, baselines can handle some attribute-related customizations well, as shown in Fig.~\ref{fig: appendix_base_cannot} (a), and our SuDe can preserve the strong ability of the baseline on these good customizations.

For the failures of baselines, they could be divided into two types:
\textbf{1)} The baseline can only generate prompt-matching images with a very low probability, as Fig.~\ref{fig: appendix_base_cannot} (b).
\textbf{2)} The baseline cannot generate prompt-matching images, as Fig.~\ref{fig: appendix_base_cannot} (c).
Our SuDe can improve both of these two cases, for example, in Fig.~\ref{fig: appendix_base_cannot} (c), 4 out of 5 generated images can match the prompt well.




\subsection{Compare with offline method} \label{sec: offline method}
Here we evaluate the offline method ELITE~\cite{wei2023elite}, which encodes a subject image to text embedding directly with an offline-trained encoder.
In the inference of ELITE, the mask annotation of the subject is needed.
We obtain these masks by Grounding DINO~\cite{liu2023grounding}.
The results are shown in Table~\ref{tab: elite}, where we see the offline method performs well in attribute alignment (BLIP-T) but poorly in subject fidelity (DINO-I).
With our SuDe, the online Dreambooth can also achieve better attribute alignment than ELITE.

\vspace{-0.5em}

\begin{table}[thbp]
  \vspace{-0.5em}
\setlength{\tabcolsep}{0.4mm}
  \centering
  \caption{Results on stable-diffusion v1.4.
  }\label{tab: elite}
  \vspace{-1.0em}
\begin{tabular}{l|cccc}
\hline
Method             & CLIP-I & DINO-I & CLIP-T & DINO-T \\ \hline
ELITE~\cite{wei2023elite}              & 68.9   & 41.5   & 28.5   & \textcolor{red}{43.2}   \\
Dreambooth~\cite{ruiz2023dreambooth}         & \textcolor{red}{77.4}   & \textcolor{red}{59.7}   & \textcolor{red}{29.0}   & 42.1   \\
Dreambooth w/ SuDe & \textbf{77.4}   & \textbf{59.9}   & \textbf{30.5}   & \textbf{45.3}   \\ \hline
\end{tabular}
\vspace{-1.0em}
\end{table}

\begin{figure*}[htbp]
\centerline{\includegraphics[scale=1.3]{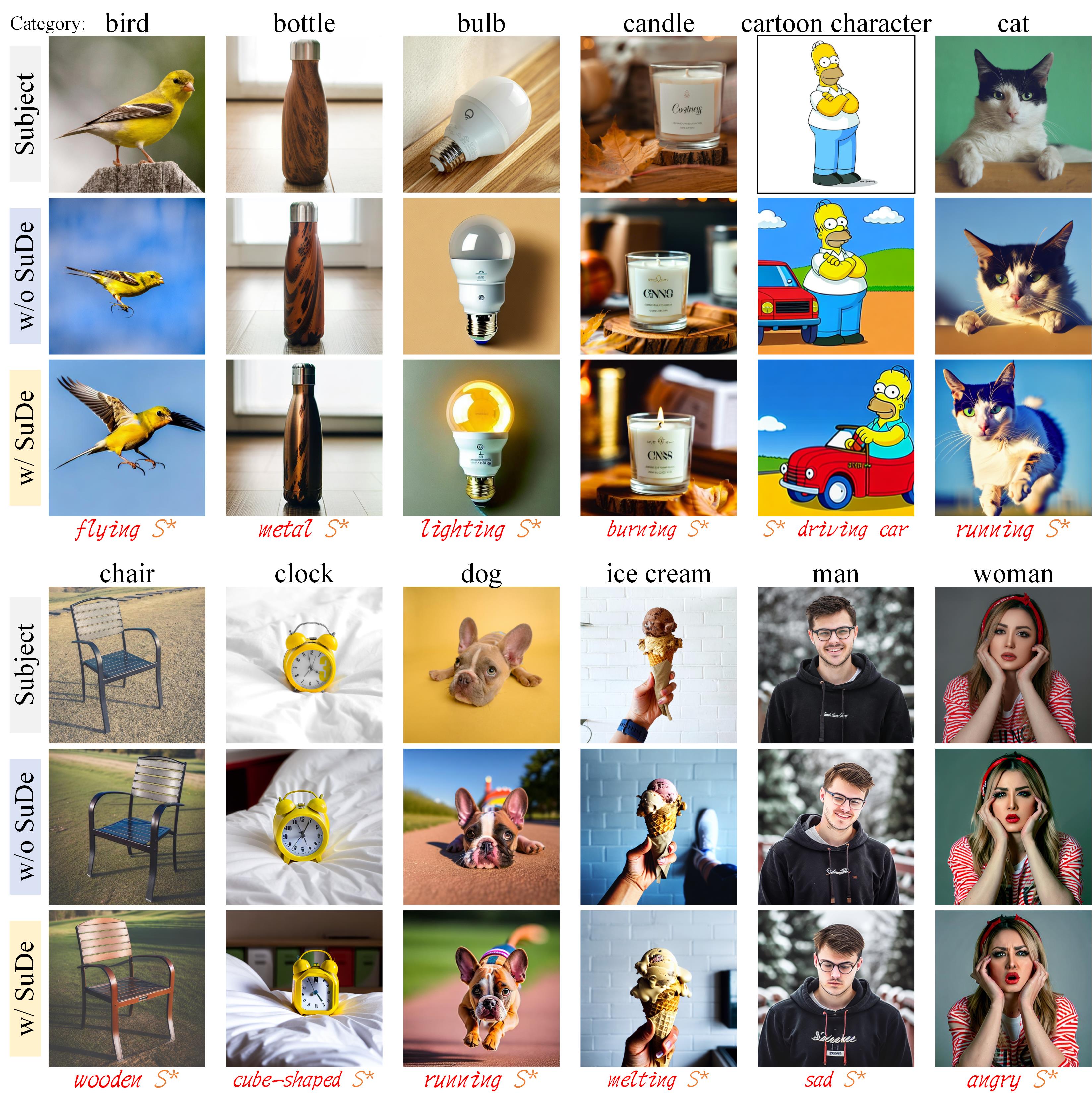}}
\caption{\textbf{More examples}. These results are obtained from DreamBooth w/o and w/ SuDe.
The subject images are from Unsplash~\cite{Unsplash}.}
\label{fig: visual result 2}
\vspace{-1.0em}
\end{figure*}

\subsection{Visualizations for more examples} \label{sec: appendix more example}
We provide more attribute-related generations in Fig.~\ref{fig: visual result 2}, where we see that based on the strong generality of the pre-trained diffusion model, our SuDe is applicable to images in various domains, such as objects, animals, cartoons, and human faces.
Besides, SuDe also works for a wide range of attributes, like material, shape, action, state, and emotion.

\subsection{Visualizations for more applications} \label{sec: appendix more vis}
\begin{figure*}[htbp]
\centerline{\includegraphics[scale=0.78]{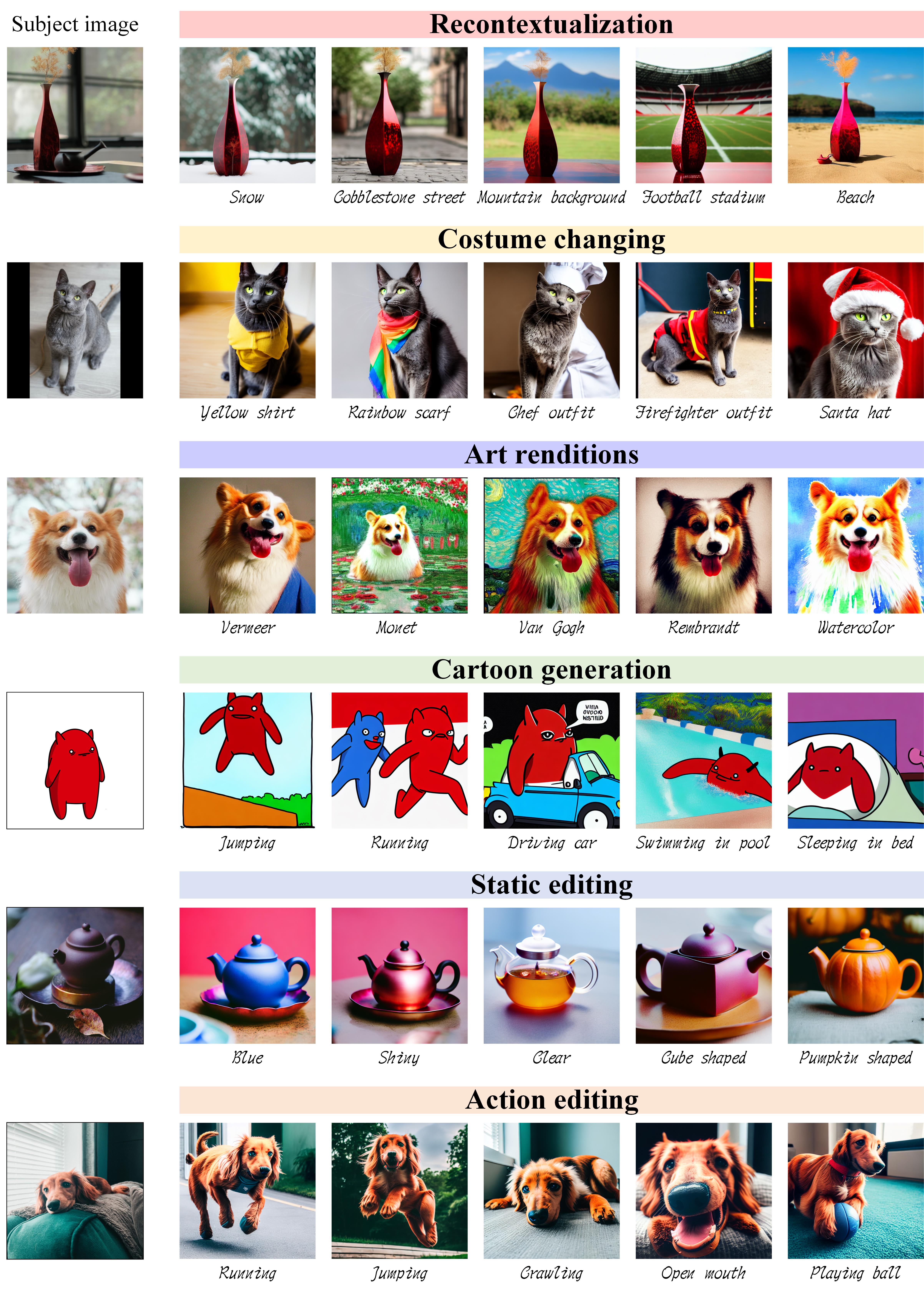}}
\caption{More applications using our SuDe with the Custom Diffusion~\cite{kumari2023multi} baseline.
}
\label{fig: appendix_more}
\end{figure*}

In Fig.~\ref{fig: appendix_more}, We present more visualization about using our SuDe in more applications, including recontextualization, art renditions, costume changing, cartoon generation, action editing, and static editing.

\end{document}